\documentclass[10pt,twocolumn,letterpaper]{article}

\usepackage[pagenumbers]{cvpr} 

\usepackage[dvipsnames,table]{xcolor}

\usepackage{graphicx}
\usepackage{booktabs}
\usepackage{multirow}
\usepackage{tikz-imagelabels}
\usepackage{graphicx}
\usepackage{caption}
\usepackage{subcaption}
\definecolor{darkerblue}{rgb}{0,0.08,0.45}
\usepackage{wrapfig}
\usepackage{adjustbox}
\usepackage{makecell}

\newcommand{\Rmnum}[1]{\uppercase\expandafter{\romannumeral #1}}  

\definecolor{color4}{rgb}{0.94,0.94,1}

\definecolor{cvprblue}{rgb}{0.21,0.49,0.74}
\usepackage[pagebackref,breaklinks,colorlinks,allcolors=cvprblue]{hyperref}

\def\paperID{***} 
\def\confName{CVPR}
\def\confYear{2025}

\title{LoRA-IR: Taming Low-Rank Experts for Efficient All-in-One Image Restoration}

\author{Yuang Ai$^{\clubsuit,\heartsuit}$ \quad Huaibo Huang$^{\clubsuit,\heartsuit\dagger}$\quad Ran He$^{\clubsuit,\heartsuit}$\\
$^{\clubsuit}$MAIS \& NLPR, Institute of Automation, Chinese Academy of Sciences \\
$^\heartsuit$School of Artificial Intelligence, University of Chinese Academy of Sciences\\
\tt\small shallowdream555@gmail.com, \tt\small huaibo.huang@cripac.ia.ac.cn, rhe@nlpr.ia.ac.cn \\ 
}

\begin{document}
\maketitle
 \newcommand\blfootnote[1]{%
\begingroup
\renewcommand\thefootnote{}\footnote{#1}%
\addtocounter{footnote}{-1}%
\endgroup
}
\blfootnote{$^{\dagger}$Corresponding author.}

\begin{abstract}
Prompt-based all-in-one image restoration (IR) frameworks have achieved remarkable performance by incorporating degradation-specific information into prompt modules. Nevertheless, handling the complex and diverse degradations encountered in real-world scenarios remains a significant challenge. To tackle this, we propose LoRA-IR, a flexible framework that dynamically leverages compact low-rank experts to facilitate efficient all-in-one image restoration. Specifically, LoRA-IR consists of two training stages: degradation-guided pre-training and parameter-efficient fine-tuning. In the pre-training stage, we enhance the pre-trained CLIP model by introducing a simple mechanism that scales it to higher resolutions, allowing us to extract robust degradation representations that adaptively guide the IR network. In the fine-tuning stage, we refine the pre-trained IR network through low-rank adaptation (LoRA). Built upon a Mixture-of-Experts (MoE) architecture, LoRA-IR dynamically integrates multiple low-rank restoration experts through a degradation-guided router. This dynamic integration mechanism significantly enhances our model's adaptability to diverse and unknown degradations in complex real-world scenarios. Extensive experiments demonstrate that LoRA-IR achieves SOTA performance across 14 IR tasks and 29 benchmarks, while maintaining computational efficiency. Code and pre-trained models will be available at: \href{https://github.com/shallowdream204/LoRA-IR}{https://github.com/shallowdream204/LoRA-IR}.
\end{abstract}    
\section{Introduction}

\begin{figure}[!t]
    \centering
    \includegraphics[width=1.0\linewidth]{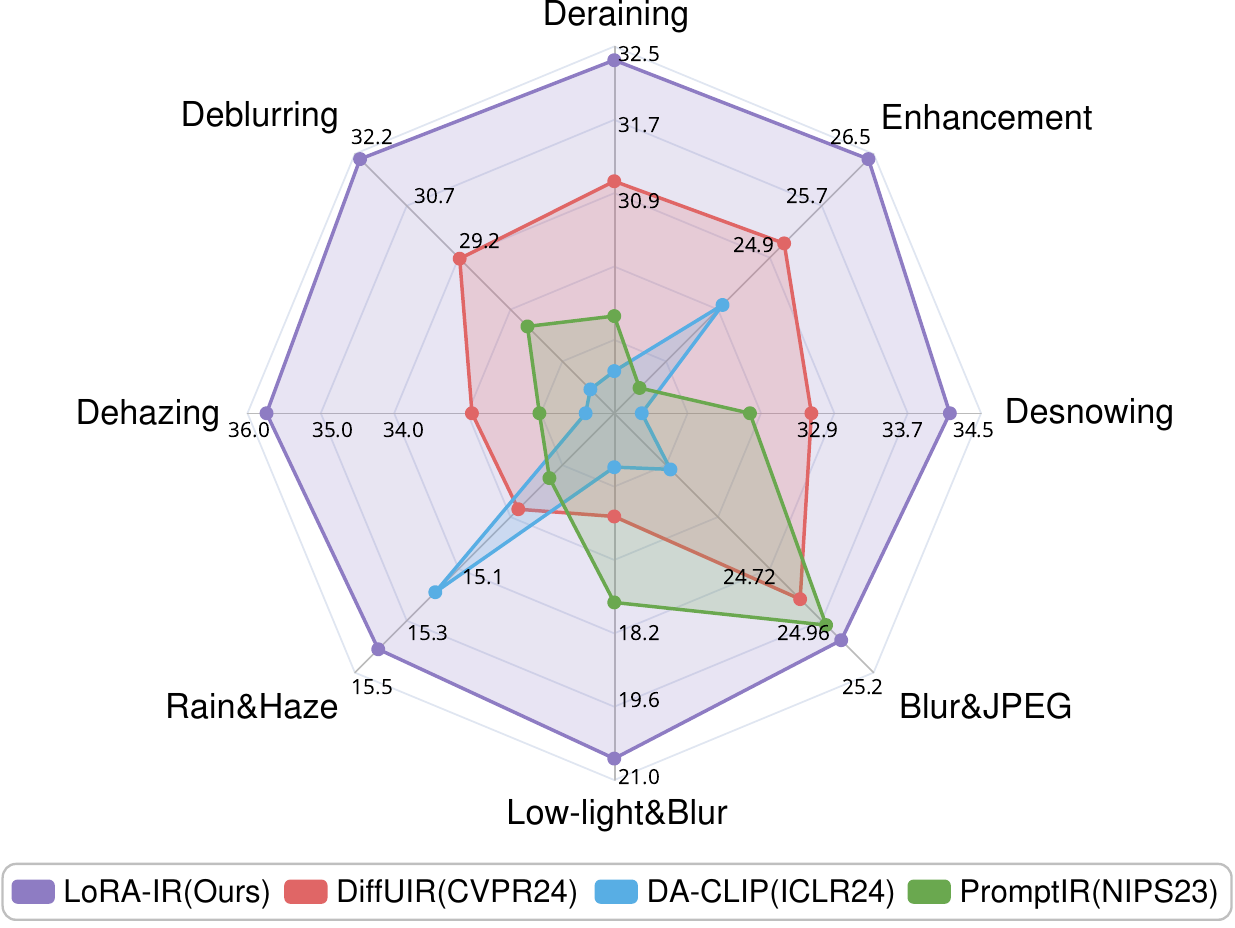}
    \vspace{-6mm}
    \caption{PSNR comparison with state-of-the-art all-in-one methods across 8 image restoration tasks (Tab.~\ref{tab:de_5} and Tab.~\ref{tab:mixed}).}
    \label{fig:psnr}
    \vspace{-0.6cm}
\end{figure}

\begin{figure*}[!t]
    \centering
    \includegraphics[width=0.96\linewidth]{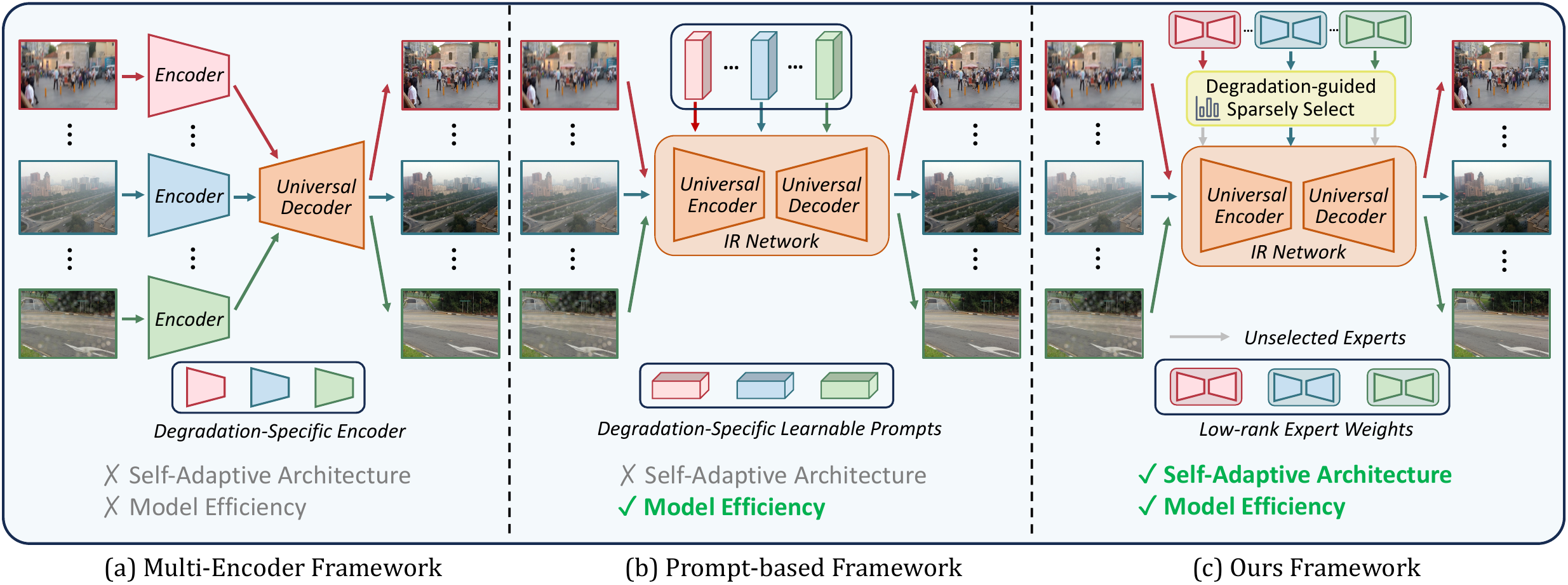}
    \captionsetup{font=small}
    \vspace{-3mm}
    \caption{
    Conceptual comparison of all-in-one frameworks. (a) Multi-Encoder Structures: Use multiple encoders to extract features, but redundancy reduces model efficiency. (b) Prompt-Based Methods: Employ lightweight prompts for degradation-specific features, improving efficiency. However, static network structures limit their ability to handle unknown complex degradations. (c) Our Proposed Framework: Self-adaptively and sparsely combines low-rank restoration experts. This design preserves model efficiency while enabling self-adaptation to various degradation types, thereby enhancing its real-world performance.
    }
    \label{fig:fig1}
    \vspace{-0.2cm}
\end{figure*}

Image restoration (IR) is a fundamental task in computer vision, aiming to recover high-quality (HQ) images from degraded low-quality (LQ) inputs. 
In recent years, significant progress has been achieved with specialized restoration networks targeting specific degradations~\cite{drsformer,zhang2017beyond,stripformer,skf}. 
However, in practical applications like autonomous driving and outdoor surveillance~\cite{mao2017can,zhu2016traffic}, images are often simultaneously affected by multiple complex degradations, including haze, rain, low-light conditions, motion blur, \etc. These intricate degradations not only degrade image quality but also severely impair the performance of downstream vision tasks, posing significant challenges to the safety and reliability of such systems. Specialized models designed for single-task restoration often struggle to generalize effectively in these unpredictable and variable environments.

To overcome the limitations of specialized IR models, there is growing interest in developing all-in-one frameworks capable of handling diverse degradations. Early approaches, such as multi-encoder architectures~\cite{as2020} (Fig.~\ref{fig:fig1} (a)), employ separate encoders for different degradation types. While effective in handling multiple degradations, their redundant structures lead to a large number of parameters, hindering scalability and efficiency. More recent state-of-the-art methods adopt prompt-based frameworks~\cite{promptir,liu2023unifying,daclip,ai2024multimodal} (Fig.~\ref{fig:fig1} (b)), encoding degradation-specific information into lightweight prompts to guide a shared network. However, relying solely on lightweight prompts and a static shared network may not fully capture the fine-grained details and specific patterns associated with different degradations, leading to suboptimal restoration results. Furthermore, potential correlations and shared features among different degradations—such as common patterns in adverse weather conditions~\cite{zhu2023learning,weatherstream}—are not extensively leveraged. Leveraging these correlations could be the key to enhancing model adaptability and effectiveness in complex real-world scenarios.

In this work, we propose LoRA-IR,  a flexible framework for efficient all-in-one image restoration (Fig.~\ref{fig:fig1} (c)). Motivated by the success of Low-Rank Adaptation (LoRA)~\cite{lora} in parameter-efficient fine-tuning, we explore the use of diverse low-rank experts to model degradation characteristics and correlations efficiently. LoRA-IR involves two training stages, both guided by the proposed Degradation-Guided Router (DG-Router). DG-Router is based on the powerful vision-language model CLIP~\cite{clip}, which has demonstrated strong representation capabilities across a wide range of high-level vision tasks~\cite{llava,coop}. However, when applied to low-level tasks, its limited input resolution inevitably leads to suboptimal performance when handling high-resolution LQ images. To this end, we introduce a simple yet effective method for scaling CLIP to high resolution. Our approach involves downsampling the image and applying a sliding window technique to capture both global and local detail representations, which are subsequently fused using lightweight MLPs. With minimal trainable parameters and a short training time, DG-Router provides robust degradation representations and probabilistic guidance for the training of LoRA-IR.

In the first stage, we use the degradation representations provided by DG-Router to guide the pre-training of the IR network. The degradation representations dynamically modulate features within the IR network through the proposed Degradation-guided Adaptive Modulator (DAM). In the second stage, we fine-tune the IR network obtained from the first stage using LoRA. Based on the Mixture-of-Experts (MoE)~\cite{moe} structure, we construct a set of low-rank restoration experts. Leveraging the probabilistic guidance of the DG-Router, we sparsely select different LoRA experts to adaptively adjust the IR network. Each expert enhances the network's ability to capture degradation-specific knowledge, while their collaboration equips the network with the capability to learn correlations between various degradations. The self-adaptive network structure enables LoRA-IR to adapt to diverse degradations and improves its generalization capabilities. As shown in Fig.~\ref{fig:psnr}, LoRA-IR outperforms all compared state-of-the-art all-in-one methods and demonstrates favorable generalizability in handling complex real-world scenarios.

The main contributions can be summarized as follows:
\begin{itemize}
    \item We propose LoRA-IR, a simple yet effective baseline for all-in-one IR. LoRA-IR leverages a novel mixture of low-rank experts structure, enhancing architectural adaptability while maintaining computational efficiency.
    \item We propose a CLIP-based Degradation-Guided Router (DG-Router) to extract robust degradation representations. DG-Router requires minimal training parameters and time, offering valuable guidance for LoRA-IR.
    \item Extensive experiments across 14 image restoration tasks and 29 benchmarks validate the SOTA performance of LoRA-IR. Notably, LoRA-IR exhibits strong generalizability to real-world scenarios, including training-unseen tasks and mixed-degradation removal.

\end{itemize}
\section{Related Work}
\noindent \textbf{Image Restoration.} Image restoration for known degradations has been extensively studied~\cite{swinir,mprnet,restormer,uformer,nafnet,zhang2022accurate,li2023efficient,diffir,focalnet}. Recently, there has been significant interest in all-in-one frameworks within the community~\cite{zhang2023ingredient,zhu2023learning,valanarasu2022transweather,conde2024high}. AirNet~\cite{airnet} is the pioneering work in all-in-one IR, utilizing contrastive learning to capture degradation information. Recent SOTA methods are mostly based on prompt learning, using lightweight prompts to encode degradation information. PromptIR~\cite{promptir} proposes a plug-and-play prompt module to guide the restoration process. DA-CLIP~\cite{daclip} utilizes a prompt learning module to incorporate degradation embeddings. MPerceiver~\cite{ai2024multimodal} introduces a multi-modal prompt learning approach to harness Stable Diffusion priors. Despite achieving promising results, most existing methods use fixed network architectures, which may limit their adaptability to cover complex real-world scenarios.

\vspace{1mm}
\noindent \textbf{Vision-Language Models.} In recent years, vision-language models (VLMs) have shown strong performance across a wide range of multi-modal and vision-only tasks~\cite{clip,openclip,llava}. Among them, CLIP~\cite{clip}, as a powerful VLM, has demonstrated impressive zero-shot and few-shot capabilities across various high-level vision tasks~\cite{wanghard,zhou2023zegclip,wang2023improving}. However, in low-level vision tasks, CLIP's capabilities have been relatively less explored. DA-CLIP~\cite{daclip} is the first to incorporate CLIP into all-in-one IR, employing a ControlNet-style~\cite{controlnet} structure and using contrastive learning with image-text pairs to fine-tune CLIP.
In this work, we focus on leveraging CLIP's visual representation capabilities to efficiently capture degradation representations. 
Compared to DA-CLIP (Tab.~\ref{tab:de_predict}), our proposed DG-Router requires $64 \times$ fewer learning parameters and $4 \times$ less training time, while achieving superior performance.

\vspace{1mm}
\noindent \textbf{Parameter-efficient Fine-tuning.} With the rise of large foundational models~\cite{clip,sam,gpt4} in modern deep learning, the community has increasingly shifted its focus towards parameter-efficient fine-tuning (PEFT) methods for effective model adaptation. Among these, prompt learning~\cite{lester2021power,coop} and Low-Rank Adaptation (LoRA)~\cite{lora} are two prominent and widely used PEFT methods. As discussed above, prompt learning has been widely applied in low-level vision tasks. LoRA posits that the weight changes during model adaptation follow a low-rank structure and incorporates trainable rank decomposition matrices into the pre-trained model. Specifically, the change matrix is re-parameterized into the product of two low-rank matrices: $W=W_0+\Delta W=W_0+sBA,$
where $W_0$ represents pre-trained weight matrix, $B\in\mathbb{R}^{m\times r}$ and $A\in\mathbb{R}^{r\times n}$ are low-rank matrices, $s=\frac{\alpha}{r}$ is the scaling factor. In this work, we first introduce LoRA into the all-in-one frameworks to facilitate efficient image restoration. 
\begin{figure}[t]
    \centering
    \captionsetup{font=small}
    \includegraphics[width=0.48\textwidth]{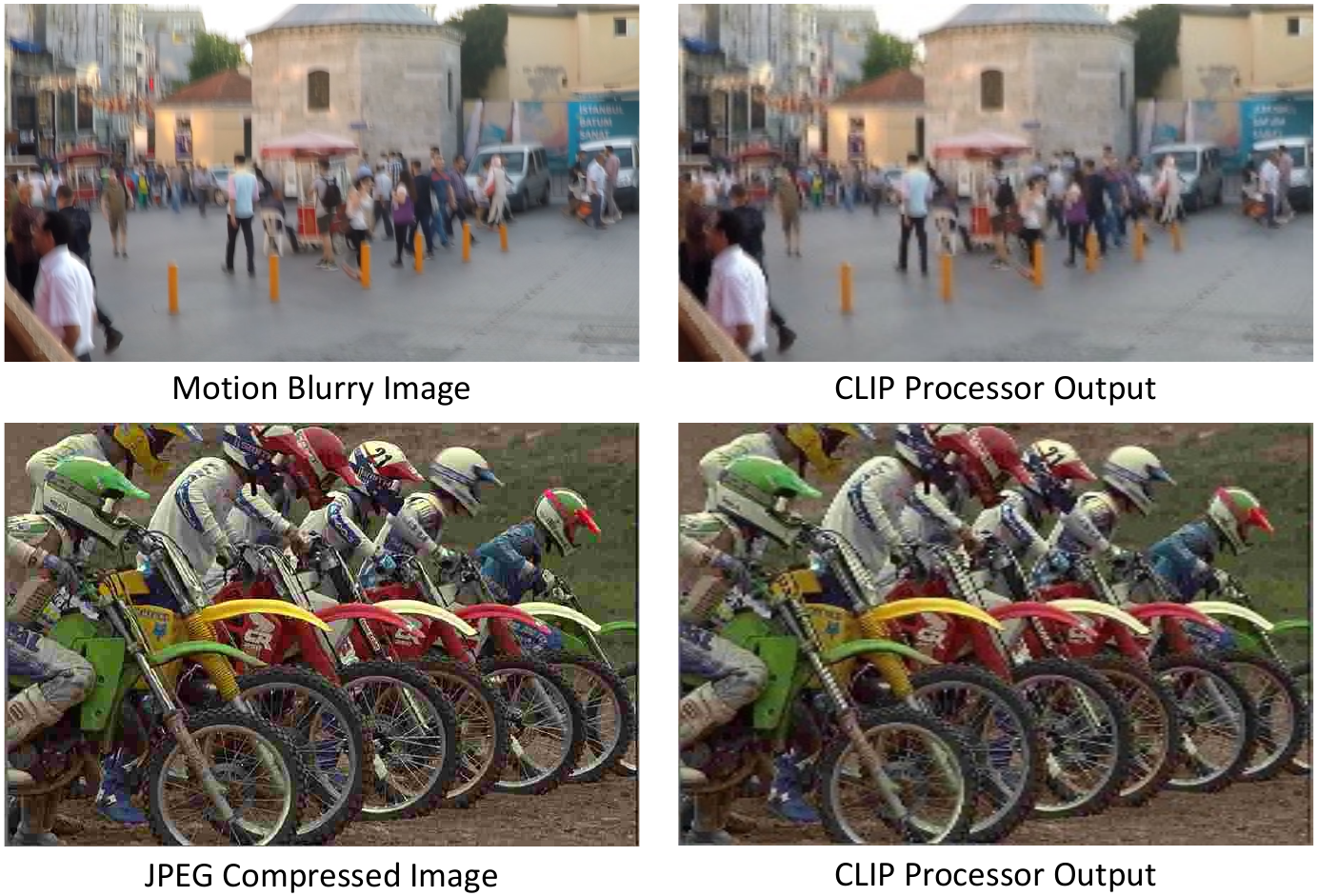} 
    \vspace{-6mm}
    \caption{Visualization of images output by the CLIP processor (top row from GoPro~\cite{gopro} and bottom row from LIVE1~\cite{sheikh2005live}), which reveals significant loss of degradation information after processing. Please zoom in for a better view.}
    \vspace{-2mm}
    \label{fig:clipdown}
\end{figure}

\begin{figure*}[t] 
\centering
    \includegraphics[width=0.98\textwidth]{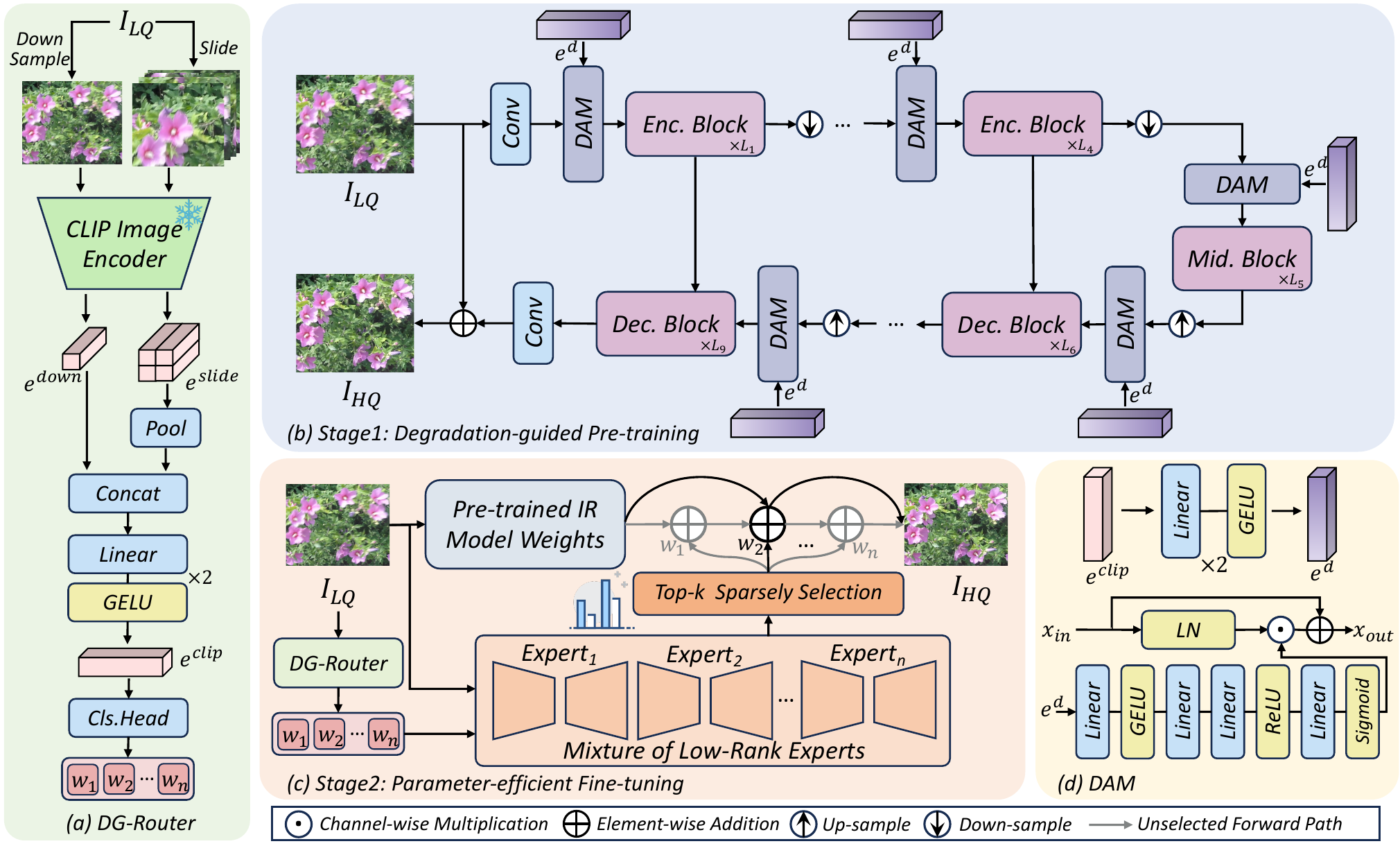}
    \vspace{-3mm}
    \caption{Overall of the proposed LoRA-IR, which includes (a) Degradation-guided router (DG-Router), (b) Pre-training image restoration network with robust degradation embedding, (c) Fine-tuning image restoration network with low-rank restoration experts, and (d) Degradation-guided adaptative modulator (DAM).} 
    \label{fig:arch}
    \vspace{-0.2cm}
\end{figure*}

\section{Method}

As shown in Fig.~\ref{fig:arch}, the image restoration network is based on the commonly used U-Net~\cite{uformer,restormer,nafnet} structure, comprising stacked encoder, middle, and decoder blocks. LoRA-IR consists of two training stages: degradation-guided pre-training and parameter-efficient fine-tuning, both guided by the proposed Degradation-Guided Router (DG-Router). Following~\cite{chen2021hinet,nafnet}, the model is optimized through PSNR loss.
We first introduce the CLIP-based DG-Router in Sec.~\ref{sec:dg_router}, which is used to extract robust degradation representations and provide probabilistic estimates to guide the training of LoRA-IR. Then we detail the pre-training process of LoRA-IR in Sec.~\ref{sec:pretrain}. Finally, we describe the fine-tuning process in Sec.~\ref{sec:finetune}.

\subsection{Degradation-guided Router}
\label{sec:dg_router}

As shown in Fig.~\ref{fig:arch} (a), DG-Router uses a pre-trained CLIP image encoder to extract rich features from LQ images. The pre-trained CLIP image encoder typically limits input images to a smaller resolution (e.g., $224\times 224$). When handling higher-resolution images, a common approach~\cite{daclip} is to downsample the image to the resolution supported by CLIP using the processor. While this may have minimal impact on perception-based high-level classification tasks, significant downsampling can potentially lead to the loss of critical degradation information in pixel-level regression tasks like image restoration. Fig.~\ref{fig:clipdown} illustrates the results after the CLIP processor has processed the LQ images. Significant downsampling causes a substantial loss of degradation information, making it difficult to effectively extract degradation representations from the CLIP output features.

To address this issue, we propose a simple yet effective mechanism for scaling up the input resolution. For input LQ image $I_{LQ} \in \mathbb{R}^{H\times W\times 3}$, we use a sliding window to partition the image into small local patches $I_{slide}\in\mathbb{R}^{M\times H_c\times W_c\times 3}$, where $M$ is the number of patches, $H_c\times W_c$ denotes the resolution supported by CLIP. Both $I_{slide}$ and down-sampled image $I_{down}\in\mathbb{R}^{H_c\times W_c\times3}$ are fed into the image encoder simultaneously, obtaining output features $e^{slide}\in\mathbb{R}^{M\times C_{clip}}$ and $e^{down}\in\mathbb{R}^{C_{clip}}$. As depicted in Fig.~\ref{fig:arch} (a), after pooling $e^{slide}$, we concatenate the features and pass them through a two-layer MLP to obtain the CLIP-extracted degradation embedding $e^{clip}$, which can be formulated as
\begin{equation}
\begin{split}
    [&e^{down}, e^{slide}] = \mathrm{CLIP}([I_{down}, I_{slide}]),\\
    &e^{clip} = \mathrm{MLP}(\mathrm{Concat}(e^{down}, \mathrm{Pooling}(e^{slide}))).
\end{split}
\end{equation}
After feeding $e^{clip}$ into the classification head, we obtain the degradation prediction probabilities $w\in \mathbb{R}^n$, where $n$ is the number of degradation types. Without bells and whistles, the DG-Router is optimized using standard cross-entropy loss, with the only parameters being the classification head and the two-layer MLP. Once training is complete, all parameters of the DG-Router are frozen and no longer updated.

\subsection{Degradation-guided Pre-training }
\label{sec:pretrain}
In the pre-training stage (Fig.~\ref{fig:arch} (b)), We dynamically modulate the restoration network using the degradation representations $e^{clip}$ extracted by the DG-Router. We propose a Degradation-guided Adaptive Modulator (DAM) to modulate the features of the restoration network. As shown in Fig.~\ref{fig:arch} (d), we first use a two-layer MLP projector to transform $e^{clip}$ into a degradation embedding $e^d$ in the feature space of the IR network. DAM adopts a structure similar to the channel attention block~\cite{channelattn}, modulating degradation information along the channel dimension, which can be formulated as 
\begin{equation}
\begin{split}
    e^{d} &= \mathrm{MLP_{shared}}(e^{clip}), \\
    x_{out} &= \mathrm{LN}(x_{in}) \odot \mathrm{Sigmoid}(\mathrm{MLP}(e^d)) + x_{in},
\end{split}
\end{equation}
where $\odot$ denotes the channel-wise multiplication, $\mathrm{MLP_{shared}}$ denotes the MLP projector shared across different blocks, $\mathrm{LN}$ denotes LayerNorm, $x_{in}$ is the original feature in the IR network, and $x_{out}$ is the feature after modulation. Through DAM modulation, the robust degradation representations from the DG-Router effectively enhance the degradation-specific knowledge of the IR network during pre-training.

\begin{table*}[!t]
\setlength{\abovecaptionskip}{5pt}
\setlength{\belowcaptionskip}{0pt}
\scriptsize
\tabcolsep=0.0mm
    \centering
    \caption{\textbf{[Setting \Rmnum{1}]} Quantitative comparisons for \textbf{\textit{4-task adverse weather removal}}. LoRA-IR surpasses recent SOTA all-in-one techniques, including MPerceiver~\cite{ai2024multimodal} and Histoformer~\cite{histo}, across all evaluated datasets and metrics.}
    \label{tab:allweather}
    \vspace{-1.5mm}
    \scalebox{1.0}{
    \begin{subtable}[t]{0.410\linewidth}
        \setlength{\tabcolsep}{5pt}
        \caption{\textbf{Image Desnowing}}\label{tab:snow}%
        
        \begin{tabular}{lcccc}
            \toprule[0.15em]
                  & \multicolumn{2}{c}{\textbf{Snow100K-S}~\cite{snow100k}} & \multicolumn{2}{c}{\textbf{Snow100K-L}~\cite{snow100k}} \\
                  & PSNR  & SSIM  & PSNR  & SSIM \\
            \midrule
            SPANet~\cite{wang2019spatial} & 29.92 & 0.8260 & 23.70 & 0.7930 \\
            JSTASR~\cite{chen2020jstasr} & 31.40 & 0.9012 & 25.32 & 0.8076 \\
            RESCAN~\cite{li2018recurrent}& 31.51 & 0.9032 & 26.08 & 0.8108 \\
            DesnowNet~\cite{snow100k} & 32.33 & 0.9500 & 27.17 & 0.8983 \\
            DDMSNet~\cite{ddmsnet} & 34.34 & 0.9445 & 28.85 & 0.8772 \\
            NAFNet~\cite{nafnet} & 34.79 & 0.9497 & 30.06 & 0.9017 \\
            Restormer~\cite{restormer} & 36.01 & 0.9579 & 30.36 & 0.9068 \\
            \midrule
            All-in-One~\cite{as2020} & -     & -     & 28.33 & 0.8820 \\
            TransWeather~\cite{valanarasu2022transweather} & 32.51 & 0.9341 & 29.31 & 0.8879 \\
            Chen~\etal\cite{chen2022learning} & 34.42 & 0.9469 & 30.22 & 0.9071 \\
            WGWSNet~\cite{zhu2023learning}  & 34.31 & 0.9460 & 30.16 & 0.9007 \\
            WeatherDiff\(_{64}\)~\cite{weatherdiff}  & 35.83 & 0.9566 & 30.09 & 0.9041 \\
            WeatherDiff\(_{128}\)~\cite{weatherdiff}  & 35.02 & 0.9516 & 29.58 & 0.8941 \\
            AWRCP~\cite{ye2023adverse}  & 36.92 & 0.9652 & 31.92 & \textbf{0.9341} \\
            MPerceiver~\cite{ai2024multimodal} & 36.23 & 0.9571 & 31.02 & 0.9164 \\
            Histoformer~\cite{histo} & \underline{37.41} & \underline{0.9656} & \underline{32.16} & 0.9261 \\
                        \rowcolor{color4}
            \textbf{LoRA-IR} & \textbf{37.89} & \textbf{0.9683} & \textbf{32.28} & \underline{0.9296} \\
            \bottomrule[0.15em]
        \end{tabular}%
    \end{subtable}
    \hfill
    \begin{subtable}[t]{0.275\linewidth}
            \setlength{\tabcolsep}{5pt}
        \caption{\textbf{Deraining \& Dehazing}}\label{tab:rainfog}%
        \begin{tabular}{lcc}
            \toprule[0.15em]
                  & \multicolumn{2}{c}{\textbf{Outdoor-Rain}~\cite{outdoor}} \\
                  & \multicolumn{1}{c}{PSNR} & \multicolumn{1}{c}{SSIM} \\
            \midrule
            CycleGAN~\cite{zhu2017cyclegan} & 17.62 & 0.6560 \\
            pix2pix~\cite{isola2017pix2pix}& 19.09 & 0.7100 \\
            HRGAN~\cite{outdoor} & 21.56 & 0.8550 \\
            PCNet~\cite{jiang2021pcnet}& 26.19 & 0.9015 \\
            MPRNet\cite{mprnet}& 28.03 & 0.9192 \\
            NAFNet~\cite{nafnet}& 29.59 & 0.9027 \\
            Restormer~\cite{restormer}& 30.03 & 0.9215 \\
            \midrule
            All-in-One~\cite{as2020} & 24.71 & 0.8980 \\
            TransWeather~\cite{valanarasu2022transweather} & 28.83 & 0.9000 \\
            Chen~\etal\cite{chen2022learning} & 29.27  & 0.9147  \\
            WGWSNet~\cite{zhu2023learning} & 29.32 & 0.9207 \\
            WeatherDiff\(_{64}\)~\cite{weatherdiff} & 29.64 & 0.9312 \\
            WeatherDiff\(_{128}\)~\cite{weatherdiff} & 29.72 & 0.9216 \\
            AWRCP~\cite{ye2023adverse}  & 31.39 & 0.9329 \\
            MPerceiver~\cite{ai2024multimodal} & 31.25 & 0.9246 \\
            Histoformer~\cite{histo} & \underline{32.08} & \underline{0.9389} \\
                        \rowcolor{color4}
            \textbf{LoRA-IR} & \textbf{32.62} & \textbf{0.9447} \\
            \bottomrule[0.15em]
        \end{tabular}%
    \end{subtable}
    \hfill
    \begin{subtable}[t]{0.240\linewidth}
            \setlength{\tabcolsep}{5pt}
        \caption{\textbf{Raindrop Removal}}\label{tab:raindrop}%
        \begin{tabular}{lcc}
            \toprule[0.15em]
                  & \multicolumn{2}{c}{\textbf{RainDrop}~\cite{raindrop}} \\
                  & \multicolumn{1}{c}{PSNR} & \multicolumn{1}{c}{SSIM} \\
            \midrule
            pix2pix~\cite{isola2017pix2pix} & 28.02 & 0.8547 \\
            DuRN~\cite{liu2019durn}  & 31.24 & 0.9259 \\
            RaindropAttn~\cite{quan2019deep}& 31.44 & 0.9263 \\
            AttentiveGAN~\cite{raindrop} & 31.59 & 0.9170 \\
            IDT~\cite{idt} & 31.87 & 0.9313 \\
            MAXIM~\cite{tu2022maxim} & 31.87 & 0.9352 \\
            Restormer~\cite{restormer}& 32.18 &   0.9408 \\
            \midrule
            All-in-One~\cite{as2020} & 31.12 & 0.9268 \\
            TransWeather~\cite{valanarasu2022transweather} & 30.17 & 0.9157 \\
            Chen~\etal\cite{chen2022learning} & 31.81 & 0.9309 \\
            WGWSNet~\cite{zhu2023learning} & 32.38 & 0.9378 \\
            WeatherDiff\(_{64}\)~\cite{weatherdiff} & 30.71 & 0.9312 \\
            WeatherDiff\(_{128}\)~\cite{weatherdiff} & 29.66 & 0.9225 \\
            AWRCP~\cite{ye2023adverse}  & 31.93 & 0.9314 \\
            MPerceiver~\cite{ai2024multimodal} & \underline{33.21} & 0.9294 \\
            Histoformer~\cite{histo} & 33.06 & \underline{0.9441} \\
                        \rowcolor{color4}
            \textbf{LoRA-IR} & \textbf{33.39} & \textbf{0.9489} \\

            \bottomrule[0.15em]
        \end{tabular}%
    \end{subtable}
    }
\end{table*}

\subsection{Parameter-efficient Fine-tuning}
\label{sec:finetune}
In the fine-tuning stage, we aim to utilize the Low-Rank Adaptation (LoRA) technique to model degradation characteristics and correlations efficiently, enhancing the model's adaptability to real-world training-unseen degradations.

As shown in Fig.~\ref{fig:arch} (c), built upon the Mixture-of-Experts (MoE) architecture, we construct a set of low-rank restoration experts. We have a total of 
$n$ low-rank experts $\{E_1,E_2,\cdots,E_n\}$, where each expert is a learnable lightweight LoRA weight from the pre-trained restoration network in the first stage, specialized in handling a specific degradation type.

For a given input LQ image, the DG-Router predicts the degradation probability $w\in \mathbb{R}^n$, which serves as the score for selecting the appropriate experts for the restoration process. We sparsely select the top-$k$ highest-scoring experts as the most relevant ones, and achieve the final restoration result through their dynamic collaboration, formulated as
\begin{equation}
\label{eq:moe}
    x_{out}=PreMod(x_{in})+\sum _{i=1}^{k}w_{\varphi (i)}^\prime E_{\varphi (i)}(x_{in}),
\end{equation}
where $PreMod$ denotes the pre-trained module in the first stage, $\varphi(i)$ denotes the index of the $i$-th selected expert, $w^\prime\in\mathbb{R}^n$ represents the result of reapplying softmax normalization to the scores of the selected top-$k$ experts (with the weights of the unselected experts set to $0$).

Note that the sparse selection mechanism in Eq.~(\ref{eq:moe}) grants LoRA-IR a self-adaptive network structure, enhancing its capacity to represent degradation-specific knowledge. The dynamic combination mechanism,  on the other hand, enables collaboration among different restoration experts, effectively capturing the commonalities and correlations across various degradations. The design of the low-rank experts ensures the efficiency of LoRA-IR, allowing it to achieve high-performance all-in-one IR in a computationally efficient manner.
\section{Experiments}

\begin{table*}[t]
  \centering
  \caption{\textbf{[Setting \Rmnum{3}]} Quantitative comparison with all-in-one models for \textbf{\textit{3-task image restoration}}. }
  \vspace{-2mm}
  \label{tab:3de}
\scalebox{0.8}{
\setlength{\tabcolsep}{13pt}
\begin{tabular}{l|ccccc|c}
    \toprule[0.15em]
    \multicolumn{1}{l|}{\multirow{2}{*}{\textbf{Method}}}  & \textbf{Dehazing} & \textbf{Deraining} &  \multicolumn{3}{c|}{\textbf{Denoising on BSD68}~\cite{bsd}} &
  \multirow{2}{*}{\textbf{Average}} \\
    \vspace{0.5pt}
     & 
  on SOTS~\cite{reside}& on Rain100L~\cite{rain100}& $\sigma = 15$ & $\sigma = 25$ & $\sigma = 50$ & \\
    \midrule
    BRDNet~\cite{tian2020image} & 23.23 / 0.895 & 27.42 / 0.895 & 32.26 / 0.898 & 29.76 / 0.836 &  26.34 / 0.836 & 27.80 / 0.843 \\
    LPNet~\cite{gao2019dynamic} & 20.84 / 0.828 & 24.88 / 0.784 &  26.47 / 0.778 & 24.77 / 0.748 & 21.26 / 0.552 & 23.64 / 0.738  \\
    FDGAN~\cite{dong2020fd} & 24.71 / 0.924 & 29.89 / 0.933 & 30.25 / 0.910 & 28.81 / 0.868 & 26.43 / 0.776 & 28.02 / 0.883 \\
    MPRNet~\cite{mprnet} & 25.28 / 0.954 & 33.57 / 0.954 & 33.54 / 0.927 & 30.89 / 0.880 & 27.56 / 0.779 & 30.17 / 0.899 \\
    DL~\cite{fan2019general} & 26.92 / 0.391 & 32.62 / 0.931 & 33.05 / 0.914 & 30.41 / 0.861 & 26.90 / 0.740 & 29.98 / 0.875 \\
        \midrule
    AirNet~\cite{airnet} & {27.94} / \underline{0.962} &{34.90} / {0.967} & {33.92} / {0.933} & {31.26} / {0.888} & {28.00} / {0.797} & {31.20} / {0.910} \\
    PromptIR~\cite{promptir} & \underline{30.58} / \textbf{0.974} & \underline{36.37} / \underline{0.972} & \underline{33.98} / \underline{0.933} & \underline{31.31} / \underline{0.888} & \underline{28.06} / \underline{0.799} & \underline{32.06} / \underline{0.913} \\
                        \rowcolor{color4}
    \textbf{LoRA-IR} & \textbf{30.68} / 0.961 & \textbf{37.75} / \textbf{0.979} & \textbf{34.06} / \textbf{0.935} & \textbf{31.42} / \textbf{0.891} & \textbf{28.18} / \textbf{0.803} & \textbf{32.42} / \textbf{0.914} \\
    \bottomrule[0.15em]
  \end{tabular}}
  \vspace{-2mm}
\end{table*}
\begin{table}[!t] 
    \centering
    \caption{\textbf{[Setting \Rmnum{2}]} Quantitative comparisons for \textbf{\textit{3-task real-world adverse weather removal}} on WeatherStream~\cite{weatherstream}.}
    \vspace{-2mm}
\scalebox{0.8}{
\setlength{\tabcolsep}{6pt}
        \begin{tabular}{ll|c|ccc|c}
        \toprule[0.15em]
        \multicolumn{2}{l|}{\textbf{Method}} & \textbf{Venue} &
        \textbf{Rain} & \textbf{Haze} & \textbf{Snow} & \textbf{Average}  \\
        \midrule
        \multicolumn{2}{l|}{MPRNet~\cite{mprnet}} &CVPR'21  & 21.50 &  21.73 &  20.74 &  21.32   \\
        \multicolumn{2}{l|}{NAFNet~\cite{nafnet}}&ECCV'22  & 23.01  & 22.20  & 22.11 & 22.44   \\
        \multicolumn{2}{l|}{Uformer~\cite{uformer}}&CVPR'22  & 22.25 & 18.81  & 20.94  & 20.67  \\
        \multicolumn{2}{l|}{Restormer~\cite{restormer}}&CVPR'22  & 23.67  & 22.90 & 22.51 & 22.86   \\
        \multicolumn{2}{l|}{GRL~\cite{li2023efficient}}&CVPR'23  & 23.75 & 22.88  & 22.59 & 23.07\\
        \midrule
        \multicolumn{2}{l|}{AirNet~\cite{airnet}}&CVPR'22 & 22.52 & 21.56 & 21.44 & 21.84   \\
        \multicolumn{2}{l|}{TUM~\cite{chen2022learning}}&CVPR'22 & 23.22 & 22.38 & 22.25& 22.62  \\
        \multicolumn{2}{l|}{Transweather~\cite{valanarasu2022transweather}}&CVPR'22 & 22.21 & 22.55  & 21.79 & 22.18 \\
        \multicolumn{2}{l|}{WGWS~\cite{zhu2023learning}}&CVPR'23 & 23.80  & 22.78  & 22.72& 23.10  \\
        \multicolumn{2}{l|}{LDR~\cite{ldr}}&CVPR'24 & \underline{24.42} & \underline{23.11}& \underline{23.12} & \underline{23.55}\\
        \rowcolor{color4}
        \multicolumn{2}{l|}{\textbf{LoRA-IR}} &-& \textbf{25.22} & \textbf{24.39} & \textbf{23.31}  & \textbf{24.31} \\
        \bottomrule[0.15em]
    \end{tabular}{}
    }
    \label{tab:weatherstream}
    \vspace{-2mm}
\end{table}

\subsection{Experimental Setup}
\textbf{Settings.}
To comprehensively evaluate our method, we conduct experiments in five different settings following previous works: \textbf{(\Rmnum{1}) \textit{4-task adverse weather removal}}~\cite{histo}, including desnowing, deraining, dehazing, and raindrop removal; \textbf{(\Rmnum{2}) \textit{3-task real-world adverse weather removal}}~\cite{ldr}, including deraining, dehazing, and desnowing; \textbf{(\Rmnum{3}) \textit{3-task image restoration}}~\cite{airnet}, including deraining, dehazing, and denoising; \textbf{(\Rmnum{4}) \textit{5-task image restoration}}~\cite{diffuir}, including deraining, low-light enhancement, desnowing, dehazing, and deblurring; \textbf{(\Rmnum{5}) \textit{10-task image restoration}}~\cite{daclip}, including deblurring, dehazing, JPEG artifact removal, low-light enhancement, denoising, raindrop removal, deraining, shadow removal, desnowing, and inpainting. For each setting, we train a single model to handle multiple types of degradation.
\vspace{1mm}

\noindent\textbf{Datasets and Metrics.}
For Setting \Rmnum{1}, we use the AllWeather~\cite{valanarasu2022transweather,histo} dataset to evaluate our method.  For Setting \Rmnum{2}, we use the WeatherStream~\cite{weatherstream} dataset to evaluate the model's performance in real-world scenarios. For Setting \Rmnum{3}, we use RESIDE~\cite{reside} for dehazing, WED~\cite{wed} and BSD~\cite{bsd} for denoising, Rain100L~\cite{rain100} for deraining. For Setting \Rmnum{4}, we use a merged dataset~\cite{restormer,diffuir} for deraining, LOL~\cite{lol}, DCIE~\cite{dice}, MEF~\cite{mef}, and NPE~\cite{npe} for low-light enhancement, Snow100K~\cite{snow100k} for desnowing, RESIDE~\cite{reside} for dehazing,  GoPro~\cite{gopro}, HIDE~\cite{hide}, RealBlur~\cite{realblur} for deblurring. For Setting \Rmnum{5}, we use the same dataset as~\cite{daclip}. Due to space limitations, further information on the training dataset, training protocols, and additional visual results are provided in the \textbf{Appendix}.

As for evaluation metrics, we adopt PSNR and SSIM as the distortion metrics, LPIPS~\cite{lpips} and FID~\cite{fid} as perceptual metrics. For benchmarks that do not include ground truth images, we use NIQE~\cite{niqe}, LOE~\cite{loe} and IL-NIQE~\cite{ilniqe} as no-reference metrics.
\vspace{1mm}

\noindent\textbf{Implementation Details.}
For the training of DG-Router, we use the Adam optimizer with a batch-size of $64\times n$, where $n$ is the number of tasks. The whole training takes 20 minutes with a fixed learning rate of $2e^{-4}$ using 8 A100 GPUs.
Our LoRA-IR follows a two-stage training process, namely pre-training and fine-tuning. For both stages, we use the AdamW optimizer with a batch-size of 64. Following~\cite{daclip,diffuir}, the training patch size is set to 256 to ensure fair comparisons. Random cropping, flipping, and rotation are used as data augmentation techniques. For the pre-training stage,  we use an initial learning rate of $1e^{-3}$, which is updated using the cosine annealing scheduler after 200000 iterations. The minimal learning rate is set to $1e^{-5}$. For fine-tuning, we use an initial learning rate of $1e^{-4}$, which decreases to $1e^{-5}$ after 100000 iterations. For the image restoration network structure, all basic blocks in Fig.~\ref{fig:arch} are the simple convolutional NAFBlocks~\cite{nafnet}, forming a simple all-in-one CNN baseline. More specific details for different settings are provided in the \textbf{Appendix}.

\subsection{Comparison with State-of-the-Arts}

\begin{figure*}[t]
	
	\centering
	
	\newcommand{\h}{0.105}
	\newcommand{\wa}{0.12}
	\newcommand{\wb}{0.16}
	\newcommand{\g}{-0.7mm}
	\setlength\tabcolsep{1.4pt}
  	\renewcommand{\arraystretch}{1}
	\resizebox{1.00\linewidth}{!} {
			\renewcommand{\h}{0.143}
			\newcommand{\w}{0.22}
				\begin{adjustbox}{valign=t}
					\begin{tabular}{cccccc}
						\includegraphics[height=\h \textwidth, width=\w \textwidth]{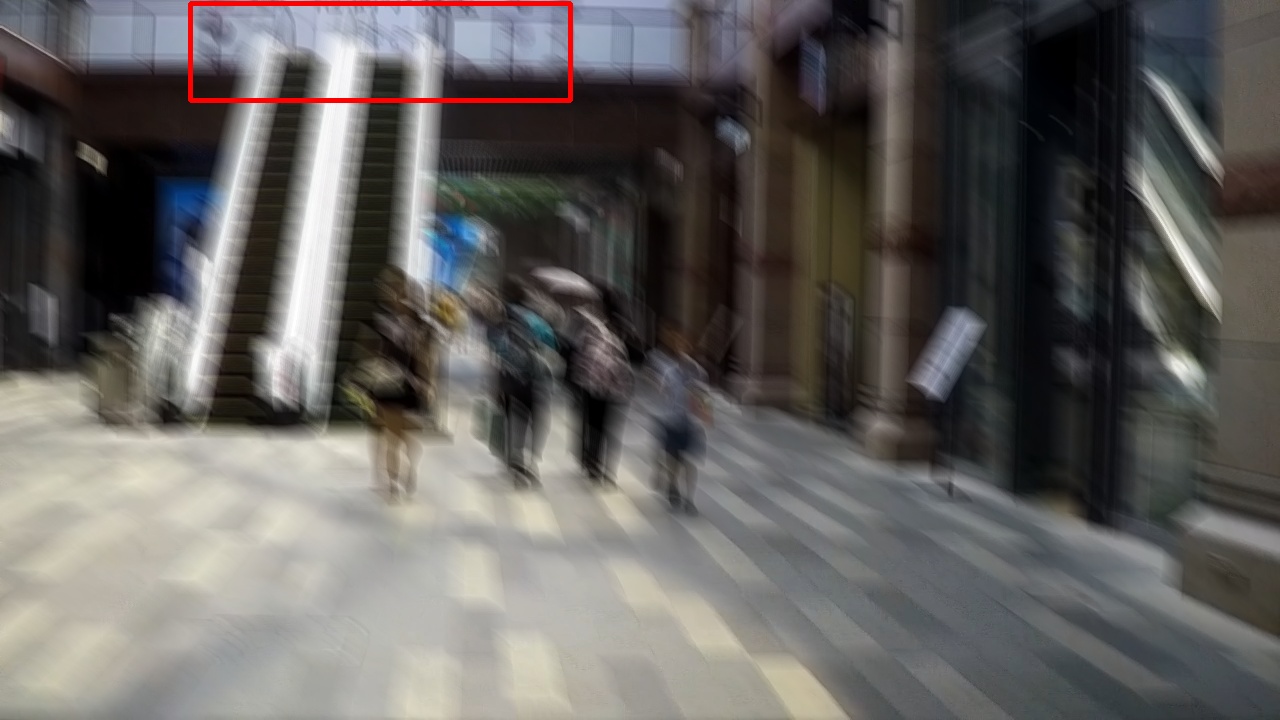} \hspace{\g} &
						\includegraphics[height=\h \textwidth, width=\w \textwidth]{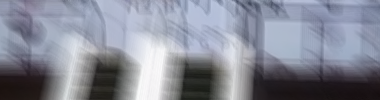} \hspace{\g} &
				\includegraphics[height=\h \textwidth, width=\w \textwidth]{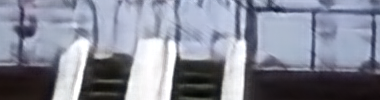} \hspace{\g} &
				\includegraphics[height=\h \textwidth, width=\w \textwidth]{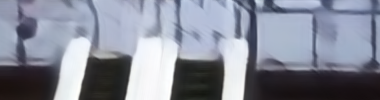} \hspace{\g} &
				\includegraphics[height=\h \textwidth, width=\w \textwidth]{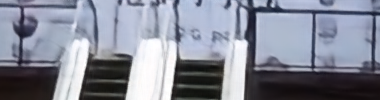} \hspace{\g} &
      			\includegraphics[height=\h \textwidth, width=\w \textwidth]{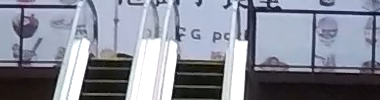} 
						\\
						 Blurry Image \hspace{\g} &
						Input \hspace{\g} &
						PromptIR~\cite{promptir}\hspace{\g} &
      					DiffUIR~\cite{diffuir} &
                 		\textbf{LoRA-IR (Ours)} &
                            GT \hspace{\g}
						\\
					\end{tabular}
				\end{adjustbox}
			
	}
  	\renewcommand{\arraystretch}{1}
	\resizebox{1.00\linewidth}{!} {
			\renewcommand{\h}{0.143}
			\newcommand{\w}{0.22}
				\begin{adjustbox}{valign=t}
					\begin{tabular}{cccccc}
						\includegraphics[height=\h \textwidth, width=\w \textwidth]{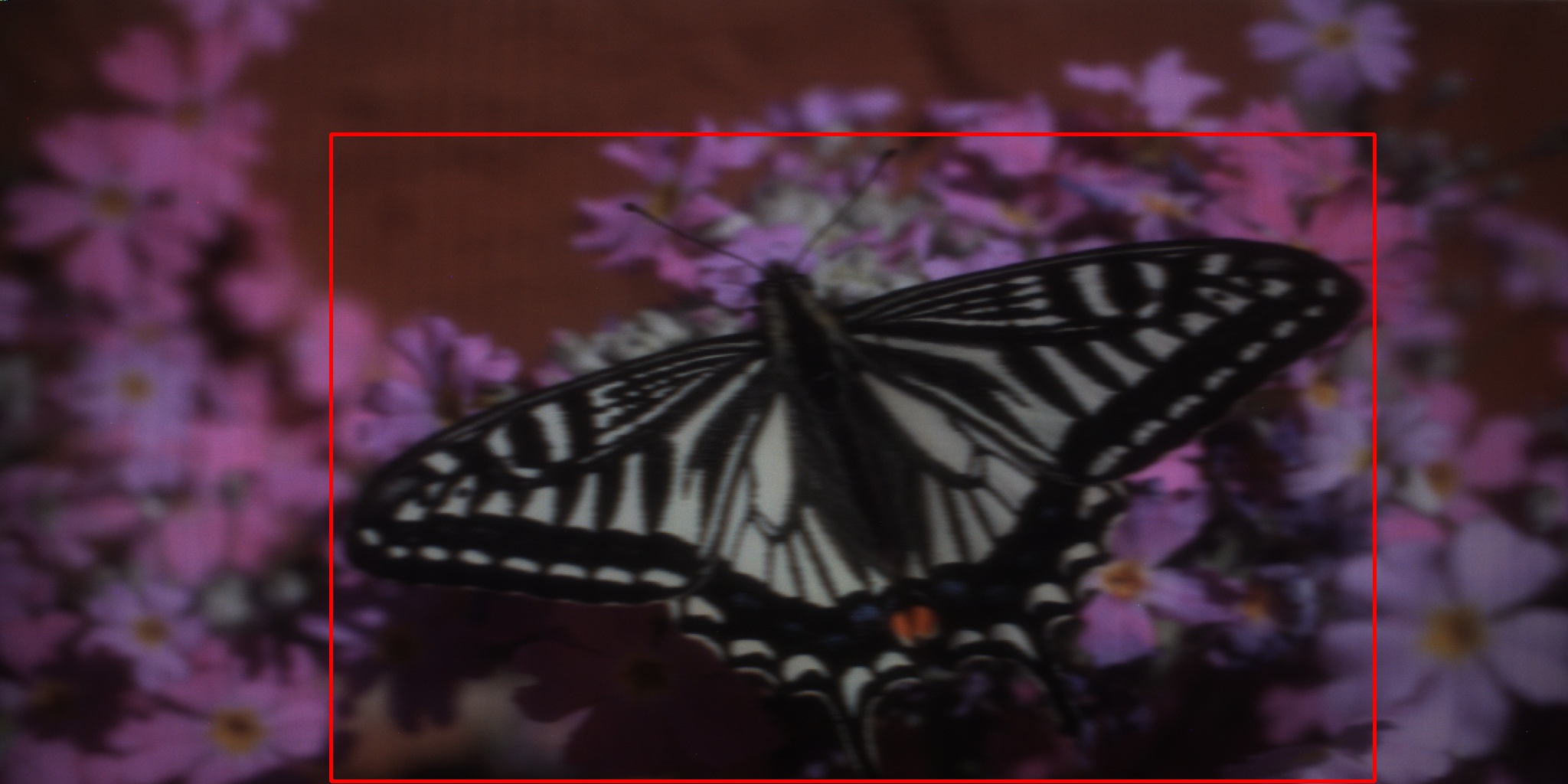} \hspace{\g} &
						\includegraphics[height=\h \textwidth, width=\w \textwidth]{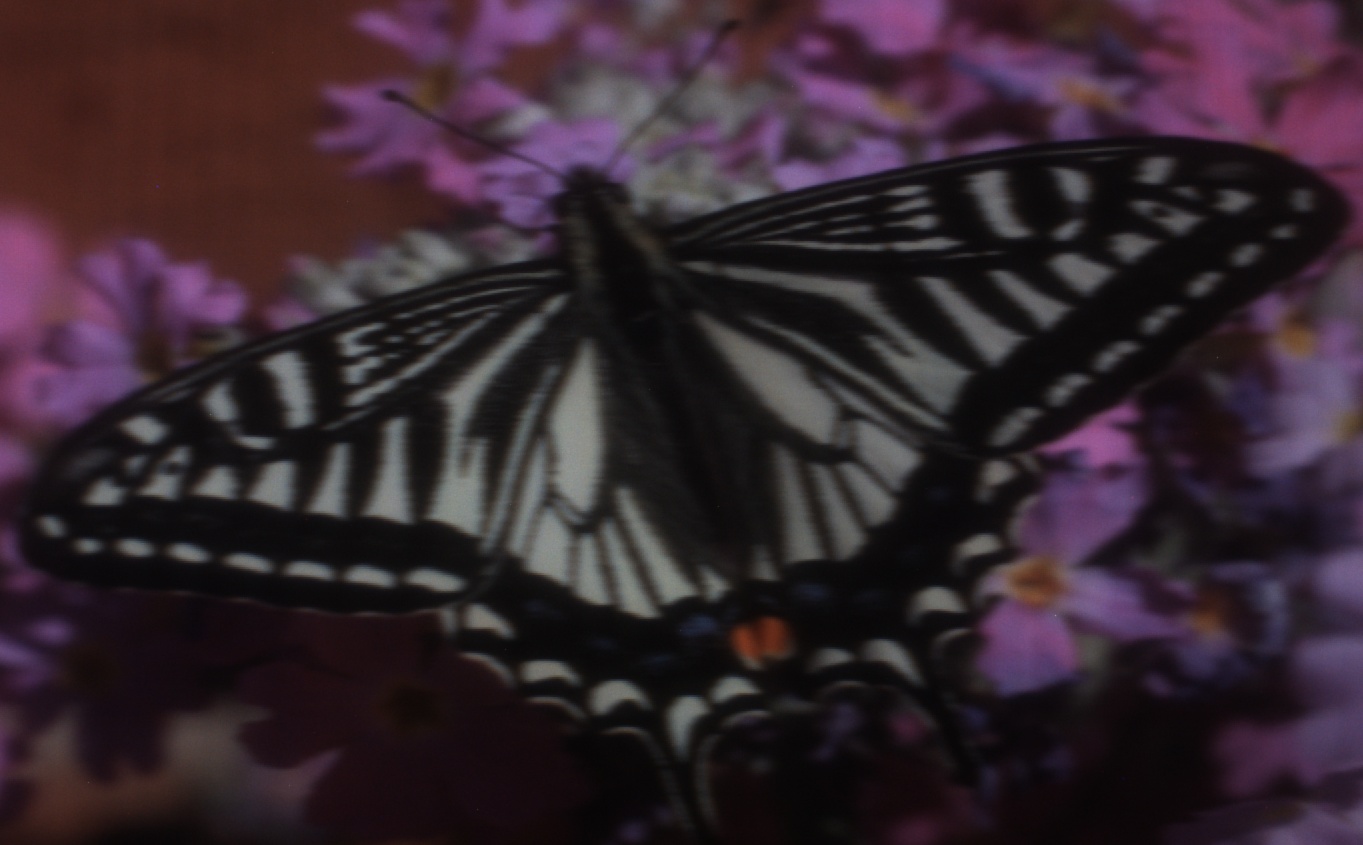} \hspace{\g} &
				\includegraphics[height=\h \textwidth, width=\w \textwidth]{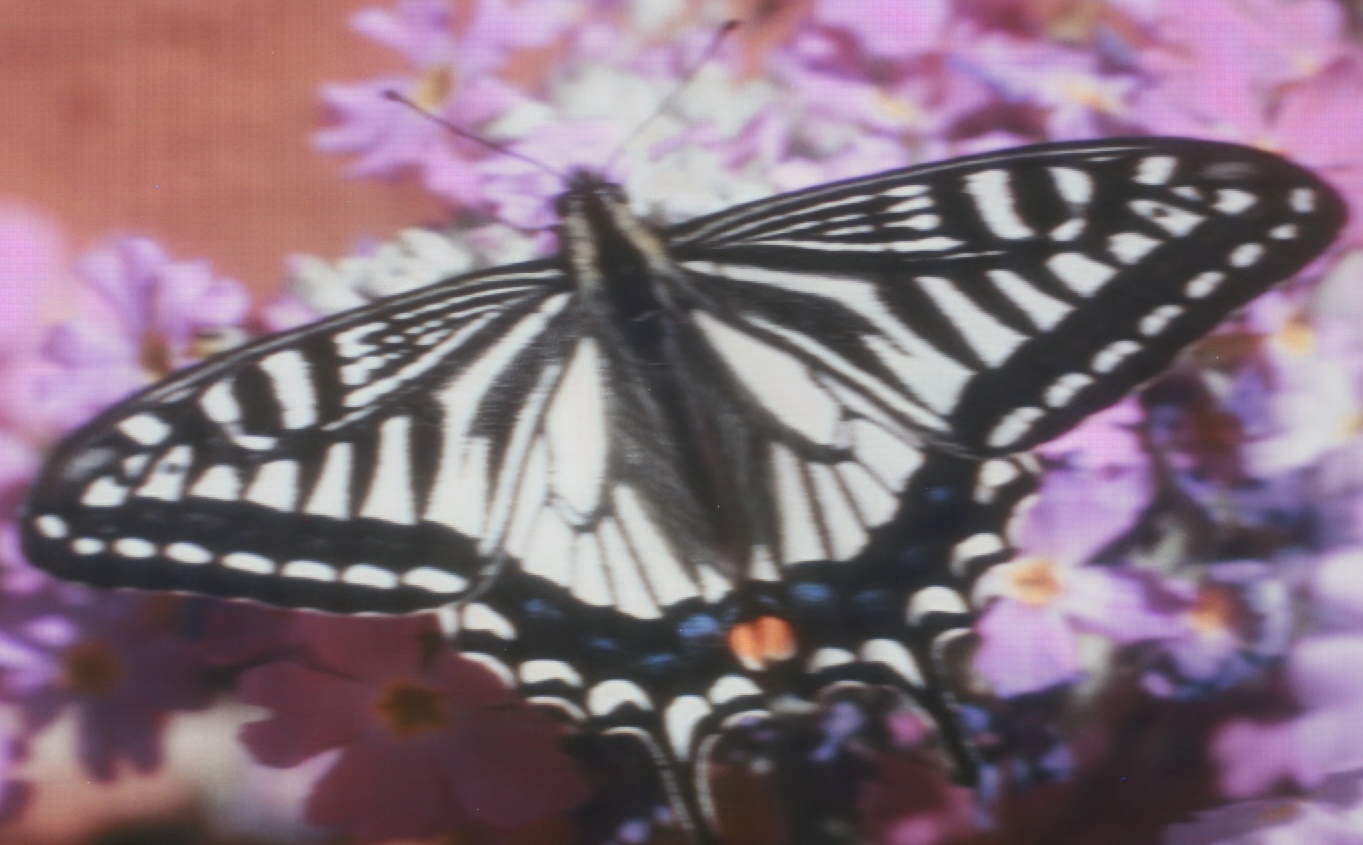} \hspace{\g} &
				\includegraphics[height=\h \textwidth, width=\w \textwidth]{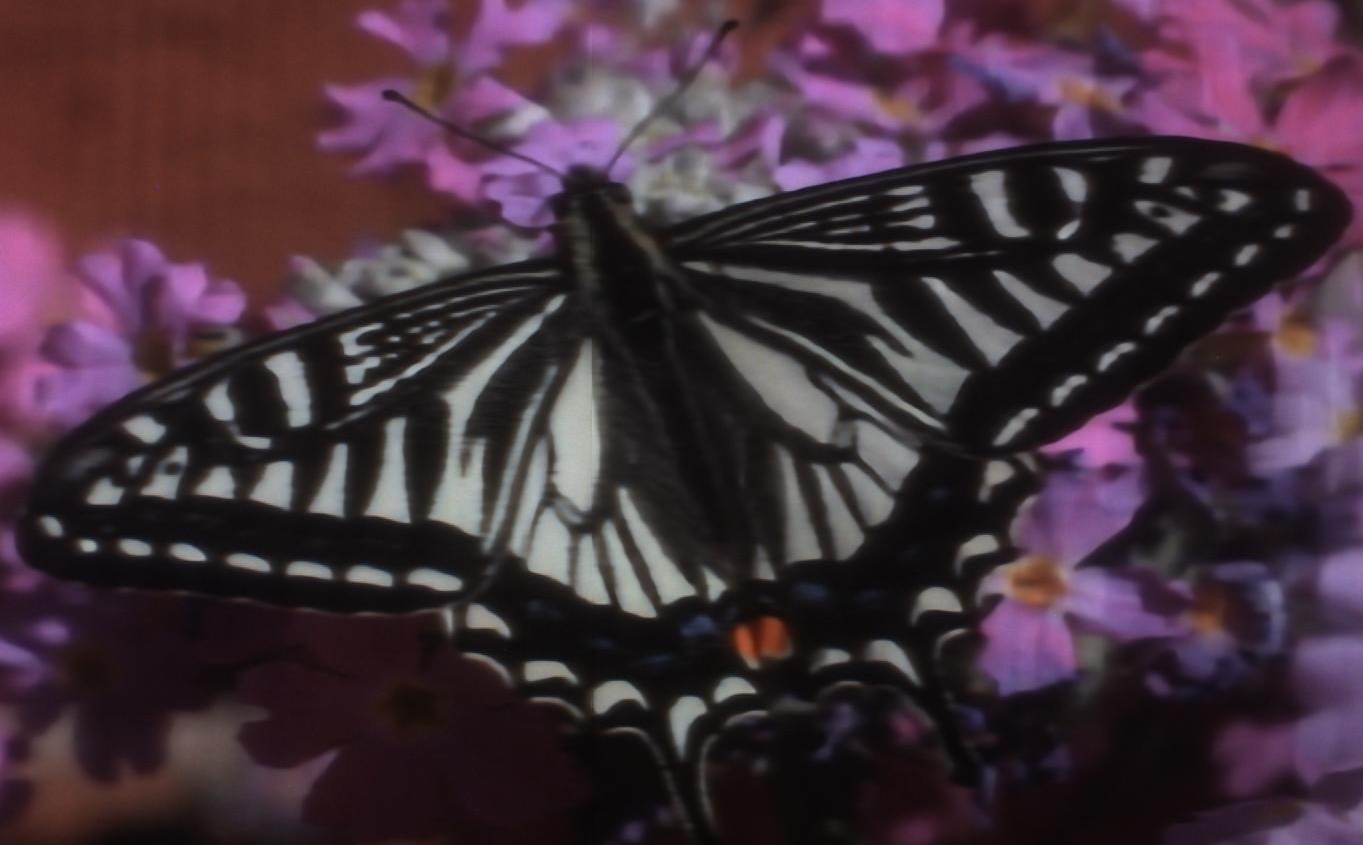} \hspace{\g} &
				\includegraphics[height=\h \textwidth, width=\w \textwidth]{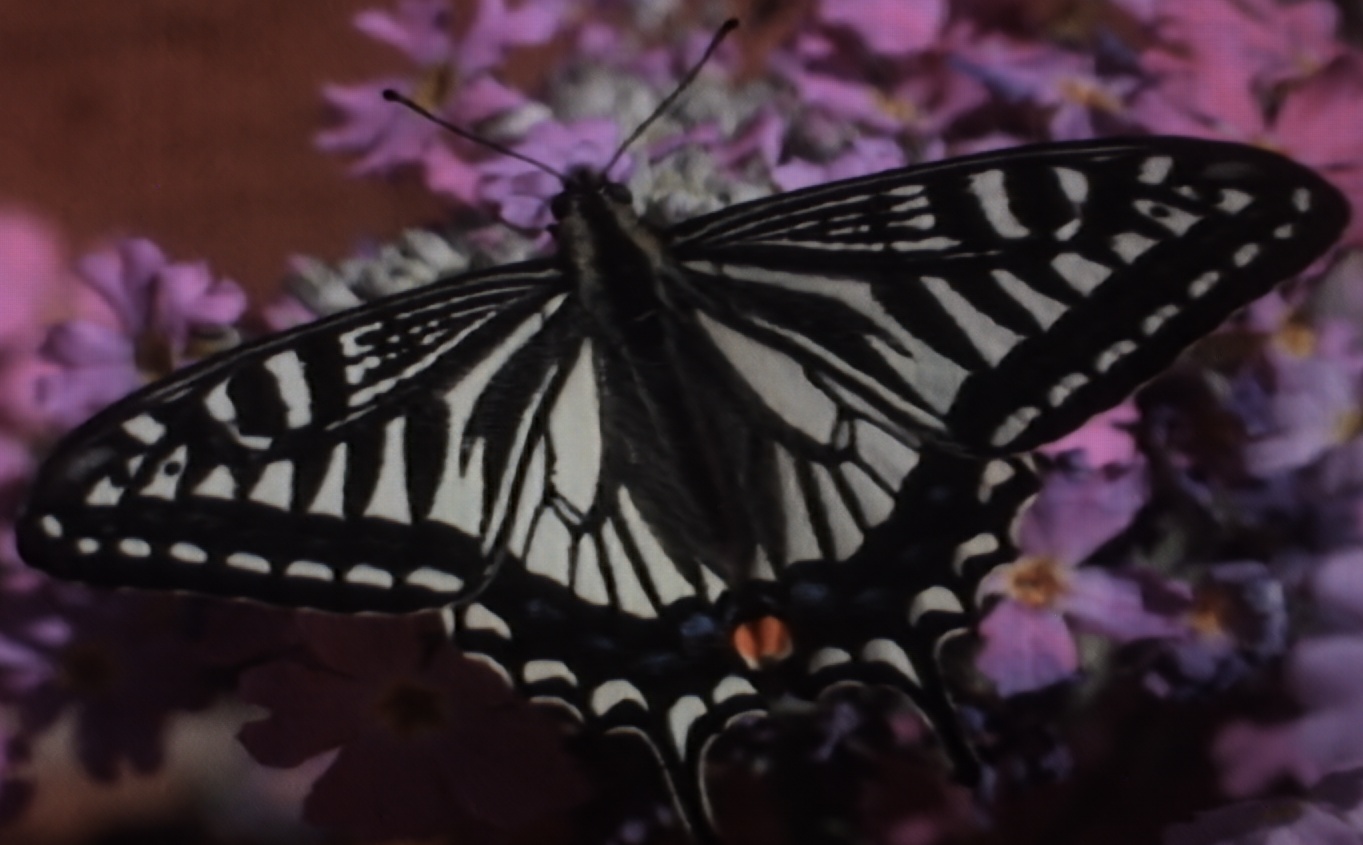} \hspace{\g} &
      			\includegraphics[height=\h \textwidth, width=\w \textwidth]{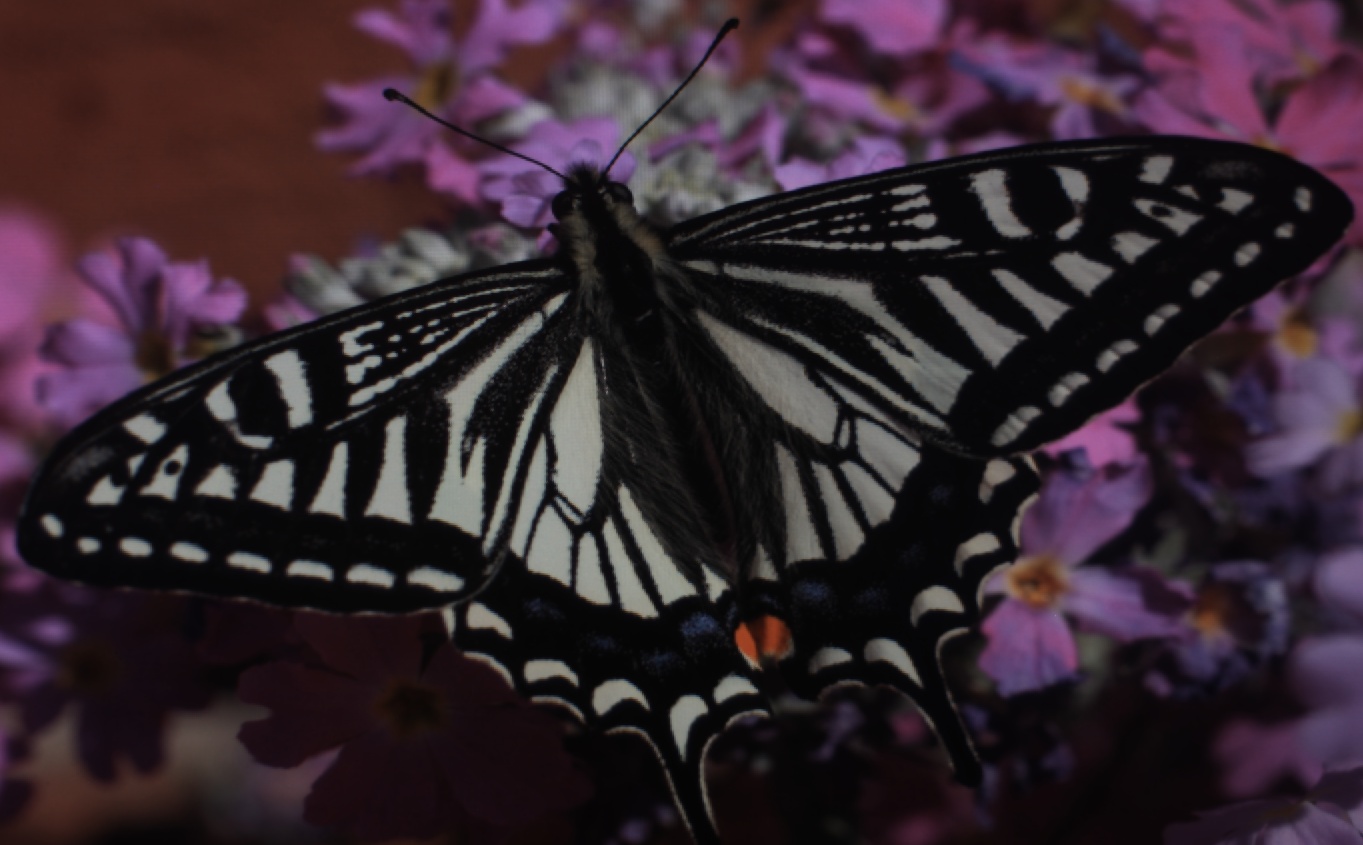} 
						\\
						 LQ Image \hspace{\g} &
						Input \hspace{\g} &
						PromptIR~\cite{promptir}\hspace{\g} &
      					DiffUIR~\cite{diffuir} &
                 		\textbf{LoRA-IR (Ours)} &
                            GT \hspace{\g}
						\\
					\end{tabular}
				\end{adjustbox}
			
	}
 \vspace{-0.3cm}
	\caption{\textbf{[Setting \Rmnum{4}]} Visual results on HIDE~\cite{hide} for \textbf{\textit{training-seen tasks generalization evaluation}} (top row) and TOLED~\cite{udc} for \textbf{\textit{training-unseen tasks generalization evaluation}} (bottom row). Zoom in for a better view.}
 \vspace{-0.3cm}
	\label{fig:visual_hide}
\end{figure*}

\begin{table*}[t]
    \centering
    \caption{\textbf{[Setting \Rmnum{4}]} Comparison with state-of-the-art task-specific and all-in-one methods for \textbf{\textit{5-task image restoration}}.}
    \vspace{-2mm}
\scalebox{0.8}{
\setlength{\tabcolsep}{6pt}
        \begin{tabular}{ll|c|cc|cc|cc|cc|cc}
        \toprule[0.15em]
        \multicolumn{2}{l|}{\multirow{2}{*}{\textbf{Method}}} &  \multirow{2}{*}{\textbf{Venue}} & 
        \multicolumn{2}{c|}{\textbf{Deraining} $(5sets)$} & \multicolumn{2}{c|}{\textbf{Enhancement}} & \multicolumn{2}{c|}{\textbf{Desnowing} $(2sets)$} & \multicolumn{2}{c|}{\textbf{Dehazing}} & \multicolumn{2}{c}{\textbf{Deblurring}} \\
        \multicolumn{2}{c|}{} & \multicolumn{1}{c|}{} & PSNR $\uparrow$ & SSIM $\uparrow$ & PSNR $\uparrow$ & SSIM $\uparrow$ & PSNR $\uparrow$ & SSIM $\uparrow$ & PSNR $\uparrow$ & SSIM $\uparrow$  & PSNR $\uparrow$ & SSIM $\uparrow$ \\
        \midrule[0.1em]
        \multicolumn{12}{l}{{Task-Specific}}\\
        \midrule[0.1em]
        \multicolumn{2}{l|}{SwinIR~\cite{swinir}}& {ICCVW'21}  & - & - & 17.81 & 0.723 & - & - & 21.50 & 0.891 & 24.52 & 0.773  \\
        \multicolumn{2}{l|}{MIRNetV2~\cite{mirnetv2}}&{TPAMI'22}  & - & - & 24.74 & 0.851 & - & - & 24.03 & 0.927 & 26.30 & 0.799  \\
        \multicolumn{2}{l|}{IR-SDE~\cite{irsde}}& {ICML'23} & - & - & 20.45 & 0.787 & - & - & - & - & \underline{30.70} & \underline{0.901}  \\
        \multicolumn{2}{l|}{WeatherDiff~\cite{weatherdiff}}& {TPAMI'23}  & - & - & - & - & \underline{33.51} & \underline{0.939} & - & - & - & -   \\
        \multicolumn{2}{l|}{RDDM~\cite{rddm}}& {CVPR'24}  & 30.74 & 0.903 & 23.22 & {0.899} & 32.55 & 0.927 & 30.78 & 0.953 & 29.53 & 0.876  \\
        \midrule[0.1em]
        \multicolumn{12}{l}{{All-in-One}}\\
        \midrule[0.1em]
        \multicolumn{2}{l|}{Restormer~\cite{restormer}}& {CVPR'22}  & 27.10 & 0.843 & 17.63 & 0.542  & 28.61 & 0.876 & 22.79 & 0.706 & 26.36 & 0.814  \\
        \multicolumn{2}{l|}{AirNet~\cite{airnet}}& {CVPR'22}  & 24.87 & 0.773 & 14.83 & 0.767 & 27.63 & 0.860 & 25.47 & 0.923 & 26.92 & 0.811 \\
        \multicolumn{2}{l|}{Painter~\cite{wang2023images}}& {CVPR'23}  & 29.49 & 0.868 & 22.40 & 0.872 & - & - & - & - & - & - \\
        \multicolumn{2}{l|}{IDR~\cite{zhang2023ingredient}}& {CVPR'23} & - & - & 21.34 & 0.826 & - & - & 25.24 & 0.943 & 27.87 & 0.846  \\
        \multicolumn{2}{l|}{ProRes~\cite{prores}}& {arXiv'23} & 30.67 & 0.891 & 22.73 & 0.877 & - & - & - & - & 27.53 & 0.851 \\
        \multicolumn{2}{l|}{PromptIR~\cite{promptir}}& {NeurIPS'23} & 29.56 & 0.888 & 22.89 & 0.847 & 31.98 & 0.924 & 32.02 & 0.952 & 27.21 & 0.817 \\
        \multicolumn{2}{l|}{DACLIP-UIR~\cite{daclip}}&{ICLR'24} & 28.96 & 0.853 & 24.17 & 0.882 & 30.80 & 0.888 & 31.39 & \underline{0.983} & 25.39 & 0.805 \\
        \multicolumn{2}{l|}{DiffUIR-L~\cite{diffuir}}& {CVPR'24} & \underline{31.03} & \underline{0.904} & \underline{25.12} & \underline{0.907} & 32.65 & 0.927 & \underline{32.94} & 0.956 & 29.17 & 0.864 \\
                            \rowcolor{color4}

        \multicolumn{2}{l|}{\textbf{LoRA-IR}}&  - & \textbf{32.35} & \textbf{0.924} & \textbf{26.42} & \textbf{0.926} & \textbf{34.16} & \textbf{0.941} & \textbf{35.74} & \textbf{0.986} & \textbf{32.05} & \textbf{0.927} \\
        \bottomrule[0.15em]
    \end{tabular}{}
    }
    \vspace{-0.4cm}
    \label{tab:de_5}
\end{table*}

\begin{table}[t]
    \centering
    \caption{\textbf{[Setting \Rmnum{4}]} Comparison on real-world benchmarks for \textbf{\textit{training-seen tasks generalization evaluation}}. Best and second best performance of all-in-one approaches are marked in \textbf{bold} and \underline{underlineded}, respectively.}
    \vspace{-2mm}
    \setlength{\tabcolsep}{1.5pt}
    \resizebox{1.0\columnwidth}{!}{
        \begin{tabular}{ll|cc|cc|cc|cc}
        \toprule[0.15em]
        \multicolumn{2}{l|}{\multirow{2}{*}{\textbf{Method}}} &
        \multicolumn{2}{c|}{\textbf{Deraining}} & \multicolumn{2}{c|}{\textbf{Enhancement}} & \multicolumn{2}{c|}{\textbf{Desnowing}} & \multicolumn{2}{c}{\textbf{Deblurring}} \\
        \multicolumn{2}{c|}{}  & NIQE$\downarrow$ & LOE$\downarrow$ & NIQE$\downarrow$ & LOE$\downarrow$ & NIQE$\downarrow$ & IL-NIQE$\downarrow$ & PSNR$\uparrow$ & SSIM$\uparrow$  \\
        \midrule[0.1em]
        \multicolumn{10}{l}{{Task-Specific}}\\
        \midrule[0.1em]
        \multicolumn{2}{l|}{WeatherDiff~\cite{weatherdiff}} & - & - & - & - & 2.96 & {21.976} & - & - \\
        \multicolumn{2}{l|}{CLIP-LIT~\cite{liang2023iterative}} & - & - & 3.70 & 232.48 & - & - & - & - \\
        \multicolumn{2}{l|}{RDDM~\cite{rddm}} & {3.34} & 41.80 & {3.57} & {202.18} & {2.76} & 22.261 & \underline{30.74} & \underline{0.894} \\
        \multicolumn{2}{l|}{Restormer~\cite{restormer}} & 3.50 & {30.32} & 3.80 & 351.61 & - & - & {32.12} & {0.926} \\
        \midrule[0.1em]
        \multicolumn{10}{l}{All-in-One}\\
        \midrule[0.1em]
        \multicolumn{2}{l|}{AirNet~\cite{airnet}} & 3.55 & 145.3 & 3.45 & 598.13 & 2.75 & \underline{21.638} & 16.78 & 0.628 \\
        \multicolumn{2}{l|}{PromptIR~\cite{promptir}} & 3.52 & 28.53 & 3.31 & 255.13 & 2.79 & 23.000 & 22.48 & 0.770 \\
        \multicolumn{2}{l|}{DACLIP-UIR~\cite{daclip}} & 3.52 & \underline{42.03} & 3.56 & 218.27 & {2.72} & \textbf{21.498} & 17.51 & 0.667 \\
        \multicolumn{2}{l|}{DiffUIR~\cite{diffuir}} & \textbf{3.38} & \textbf{24.82} & \textbf{3.14} & \underline{193.40} & 2.74 & 22.426 & {30.63} & {0.890} \\
        \rowcolor{color4}
        \multicolumn{2}{l|}{\textbf{LoRA-IR}} & 
    \underline{3.47} & {67.53} & \underline{3.28} & \textbf{93.32} & \textbf{2.70} & 22.010 & \textbf{30.80} & \textbf{0.907} \\
        \bottomrule[0.15em]
    \end{tabular}{}
    }
    \vspace{-3mm}
    \label{tab:de5_real}
\end{table}

\begin{table}[t]
    \centering
    \caption{\textbf{[Setting \Rmnum{4}]} Comparison on TOLED and POLED datasets~\cite{udc} for \textbf{\textit{training-unseen tasks generalization evaluation}} (under-display camera image restoration).}
    \vspace{-2mm}
    \resizebox{0.96\columnwidth}{!}{
        \begin{tabular}{ll|ccc|ccc}
        \toprule[0.15em]
        \multicolumn{2}{l|}{\multirow{2}{*}{\textbf{Method}}} &
        \multicolumn{3}{c|}{\textbf{TOLED}~\cite{udc}} & \multicolumn{3}{c}{\textbf{POLED}~\cite{udc}} \\
        \multicolumn{2}{c|}{}  & PSNR $\uparrow$ & SSIM $\uparrow$ & LPIPS $\downarrow$ & PSNR $\uparrow$ & SSIM $\uparrow$ & LPIPS $\downarrow$  \\
        \midrule        
        \multicolumn{2}{l|}{NAFNet~\cite{nafnet}} & {26.89} & {0.774} & {0.346} & 10.83 & 0.416 & 0.794 \\
        \multicolumn{2}{l|}{HINet~\cite{chen2021hinet}} & 13.84 & 0.559 & 0.448 & 11.52 & {0.436} & 0.831 \\
        \multicolumn{2}{l|}{MPRNet~\cite{mprnet}} & 24.69 & 0.707 & 0.347 & 8.34 & 0.365 & 0.798 \\
        \multicolumn{2}{l|}{DGUNet~\cite{mou2022deep}} & 19.67 & 0.627 & 0.384 & 8.88 & 0.391 & 0.810 \\
        \multicolumn{2}{l|}{MIRNetV2~\cite{mirnetv2}} & 21.86 & 0.620 & 0.408 & 10.27 & 0.425 & 0.722\\
        \multicolumn{2}{l|}{SwinIR~\cite{swinir}} & 17.72 & 0.661 & 0.419 & 6.89 & 0.301 & 0.852 \\
        \multicolumn{2}{l|}{RDDM~\cite{rddm}} & 23.48 & 0.639 & 0.383 & {15.58} & 0.398 & {0.544} \\
        \multicolumn{2}{l|}{Restormer~\cite{restormer}} & 20.98 & 0.632 & 0.360 & 9.04 & 0.399 & 0.742 \\
        \midrule
        \multicolumn{2}{l|}{DL~\cite{fan2019general}} & 21.23 & 0.656 & 0.434 & 13.92 & 0.449 & 0.756 \\
        \multicolumn{2}{l|}{Transweather~\cite{valanarasu2022transweather}} & 25.02 & 0.718 & 0.356 & 10.46 & 0.422 & 0.760 \\
        \multicolumn{2}{l|}{TAPE~\cite{liu2022tape}} & 17.61 & 0.583 & 0.520 & 7.90 & 0.219 & 0.799 \\
        \multicolumn{2}{l|}{AirNet~\cite{airnet}} & 14.58 & 0.609 & 0.445 & 7.53 & 0.350 & 0.820 \\
        \multicolumn{2}{l|}{IDR~\cite{zhang2023ingredient}} & 27.91 & 0.795 & 0.312 & \underline{16.71} & 0.497 & 0.716 \\
        \multicolumn{2}{l|}{PromptIR~\cite{promptir}} & 16.70 & 0.688 & 0.422 & 13.16 & \underline{0.583} & 0.619 \\
        \multicolumn{2}{l|}{DACLIP-UIR~\cite{daclip}} & 15.74 & 0.606 & 0.472 & 14.91 & 0.475 & 0.739 \\
        \multicolumn{2}{l|}{DiffUIR-L~\cite{diffuir}}  & \textbf{29.55} & \textbf{0.887} & \underline{0.281} & 15.62 & 0.424 & \textbf{0.505}  \\
        \rowcolor{color4}
        \multicolumn{2}{l|}{\textbf{LoRA-IR}}  & \underline{28.68} & \underline{0.876} & \textbf{0.279} & \textbf{17.02}& \textbf{0.700} & \underline{0.600}  \\
        \bottomrule[0.15em]
    \end{tabular}{}
    }
    \vspace{-1mm}
    \label{tab:udc}
\end{table}

\textbf{Setting \Rmnum{1}.} Tab.~\ref{tab:allweather} shows the comparison results with task-specific methods and all-in-one methods. Compared to SOTA methods like MPerceiver~\cite{ai2024multimodal} and Histoformer~\cite{histo}, our approach shows significant improvements across all benchmarks and metrics.
\vspace{1mm}

\noindent \textbf{Setting \Rmnum{2}.} To further demonstrate the effectiveness of our method in mitigating real-world adverse weather conditions, we evaluate its performance on the WeatherStream~\cite{weatherstream} dataset. Tab.~\ref{tab:weatherstream} presents the quantitative comparison results of PSNR with SOTA general IR as well as all-in-one IR methods. Compared to the SOTA method LDR~\cite{ldr}, our method achieves an average PSNR improvement of 0.76 dB across the three tasks.
\vspace{1mm}

\noindent \textbf{Setting \Rmnum{3}.} Tab.~\ref{tab:3de} presents the quantitative comparison results for 3-task image restoration. Our method surpasses PromptIR~\cite{promptir} by 1.38 dB in PSNR on the Rain100L dataset, with an average improvement of 0.36 dB across the three tasks.
\vspace{1mm}

\noindent \textbf{Setting \Rmnum{4}.} Tab.~\ref{tab:de_5} presents the quantitative comparison results of our method against SOTA task-specific methods and all-in-one methods across five tasks. It shows that our method outperforms the compared all-in-one and task-specific methods across all tasks. For example, compared to the recent SOTA all-in-one method DiffUIR~\cite{diffuir}, LoRA-IR brings a PSNR improvement ranging from 0.92 dB to 2.87 dB across various tasks.

To further validate the generalizability of our method for complex degradations in real-world scenarios, we evaluate from two perspectives: 

\textbf{(\textit{1) Generalization on Training-seen Tasks:}} 
We directly test the trained all-in-one model on real-world benchmarks that were not seen during training. As shown in Tab.~\ref{tab:de5_real}, our method achieves the best PSNR and SSIM metrics for deblurring. As discussed in ~\cite{supir,seesr}, diffusion-based IR methods typically have an advantage in no-reference metrics like NIQE. However, our CNN-based model achieves comparable or even better performance on no-reference metrics compared to two SOTA diffusion-based methods, DACLIP-UIR~\cite{daclip} and DiffUIR~\cite{diffuir}. Notably, our method shows approximately a 100-point improvement in LOE performance over DiffUIR in enhancement. Fig.~\ref{fig:visual_hide} also shows that our method achieves more pleasing visual results.

\textbf{\textit{(2) Generalization on Training-unseen Tasks:}} We directly test all-in-one models on the training-unseen under-display camera image restoration task. Tab.~\ref{tab:udc} shows that, compared to general IR and all-in-one methods, our method achieves either the best or second-best performance across all metrics. Fig.~\ref{fig:visual_hide} shows that our method produces the clearest result when handling unknown degradations.
\vspace{1mm}

\begin{figure*}[!h]
	
	\centering
	
	\newcommand{\h}{0.105}
	\newcommand{\wa}{0.12}
	\newcommand{\wb}{0.16}
	\newcommand{\g}{-0.7mm}
	\setlength\tabcolsep{1.8pt}
  	\renewcommand{\arraystretch}{1}
	\resizebox{1.00\linewidth}{!} {
			\renewcommand{\h}{0.143}
			\newcommand{\w}{0.22}
				\begin{adjustbox}{valign=t}
					\begin{tabular}{cccccc}
						\includegraphics[height=\h \textwidth, width=\w \textwidth]{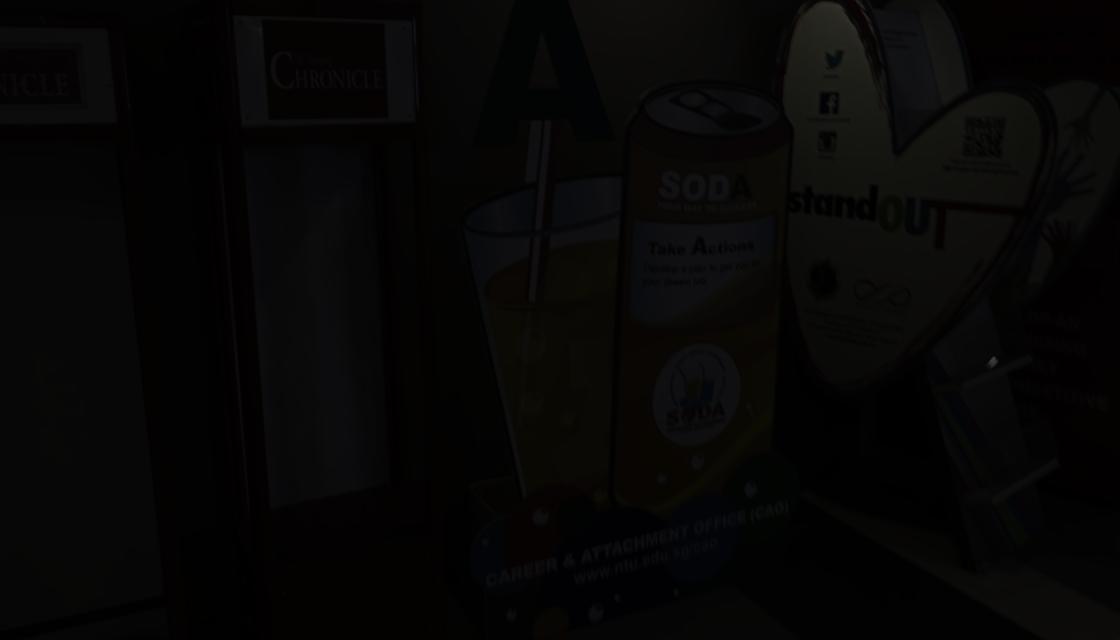} \hspace{\g} &
						\includegraphics[height=\h \textwidth, width=\w \textwidth]{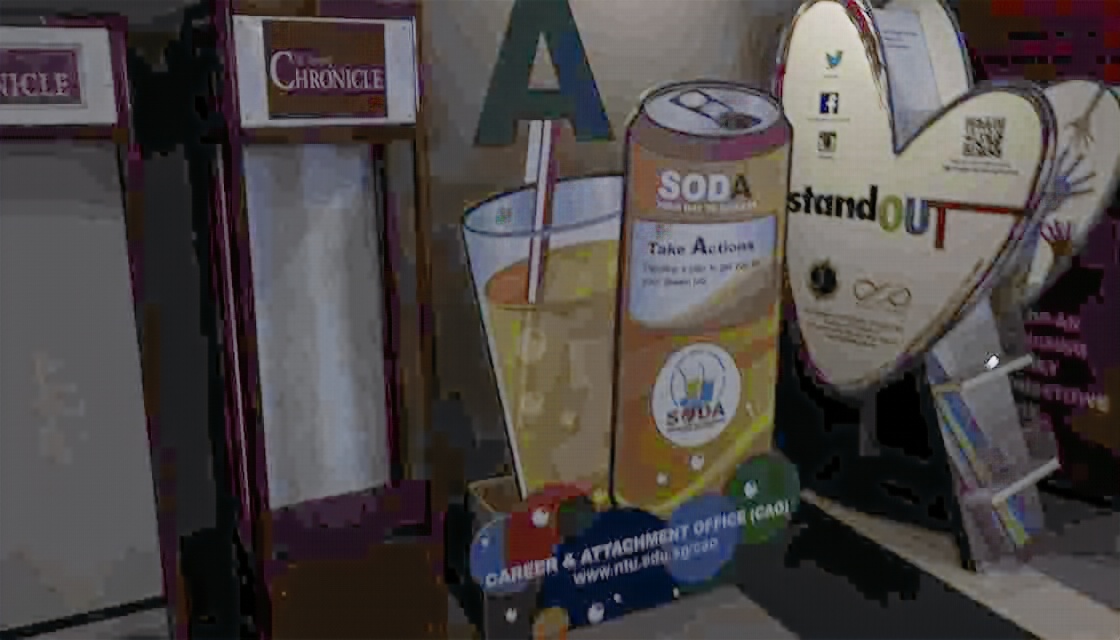} \hspace{\g} &
						\includegraphics[height=\h \textwidth, width=\w \textwidth]{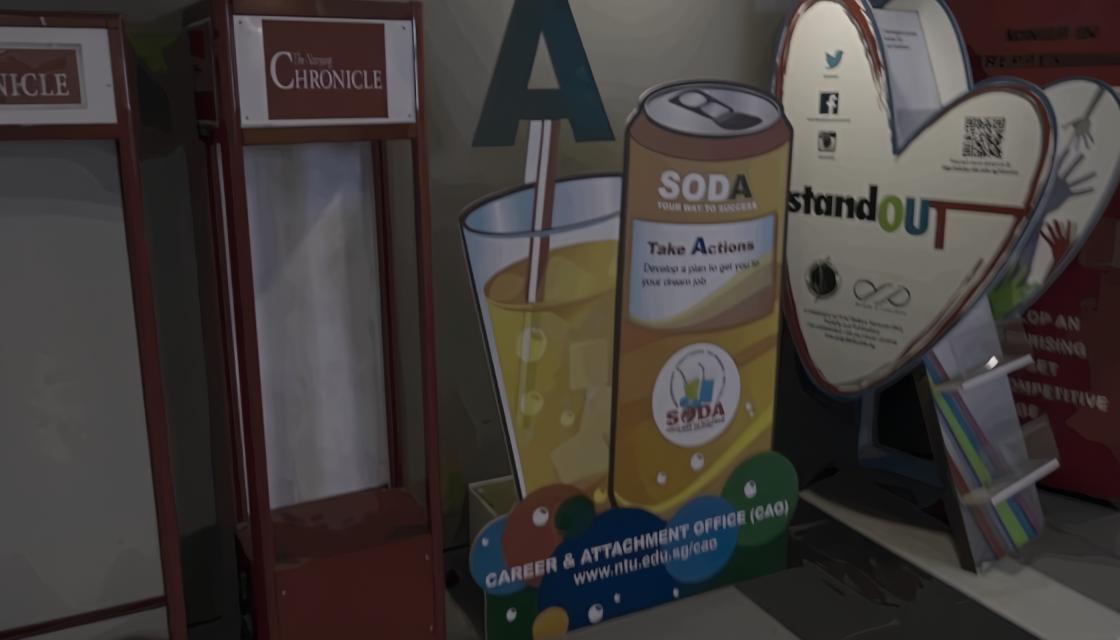} \hspace{\g} &
				\includegraphics[height=\h \textwidth, width=\w \textwidth]{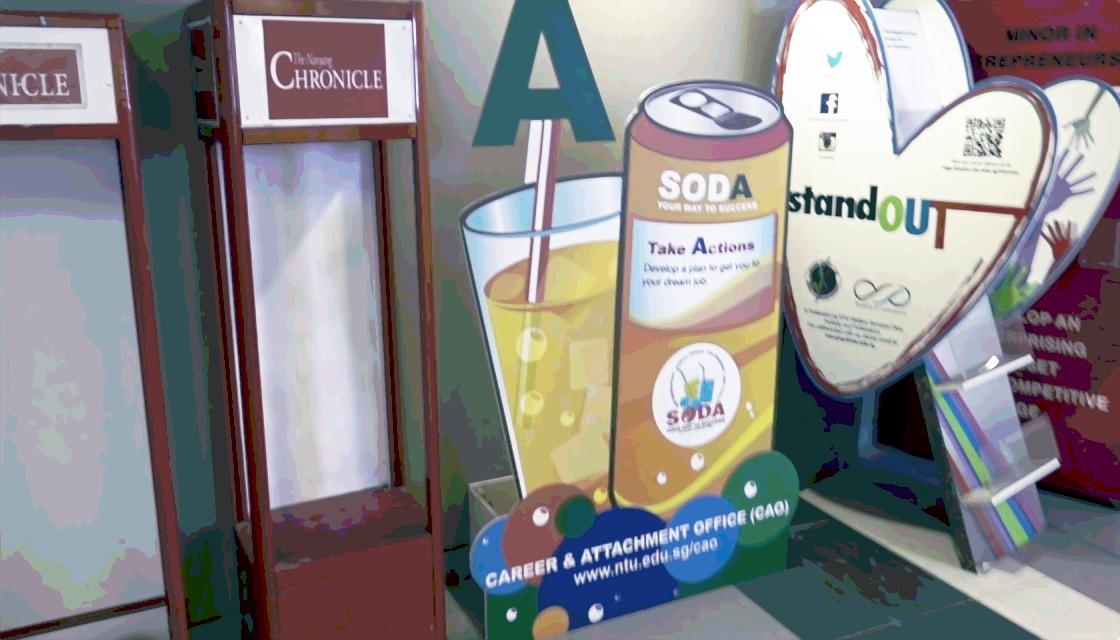} \hspace{\g} &
				\includegraphics[height=\h \textwidth, width=\w \textwidth]{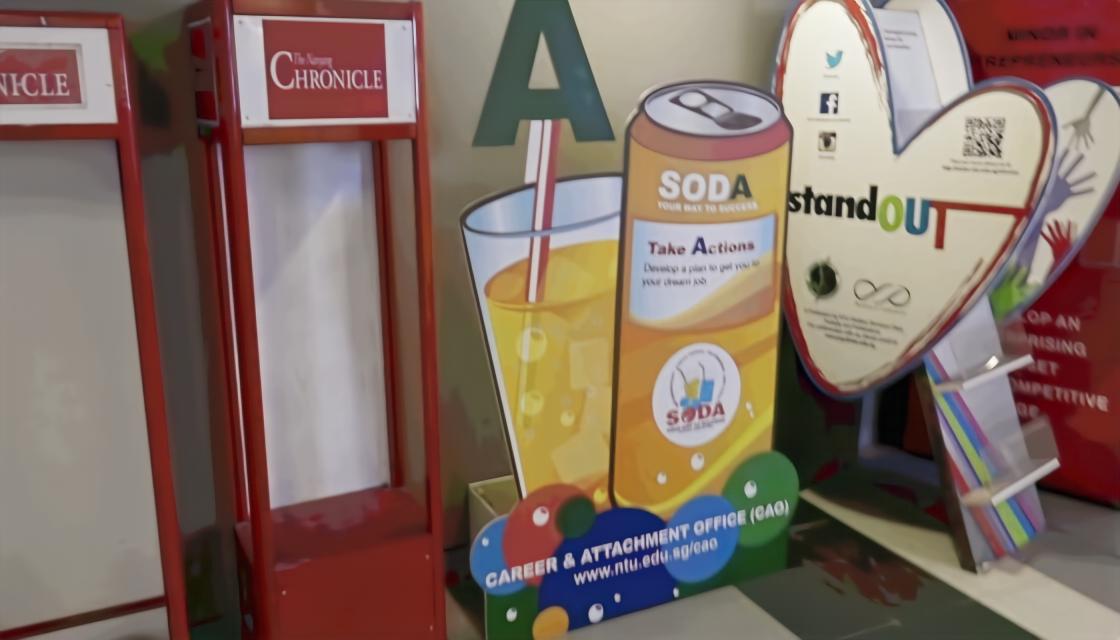} \hspace{\g} &
      			\includegraphics[height=\h \textwidth, width=\w \textwidth]{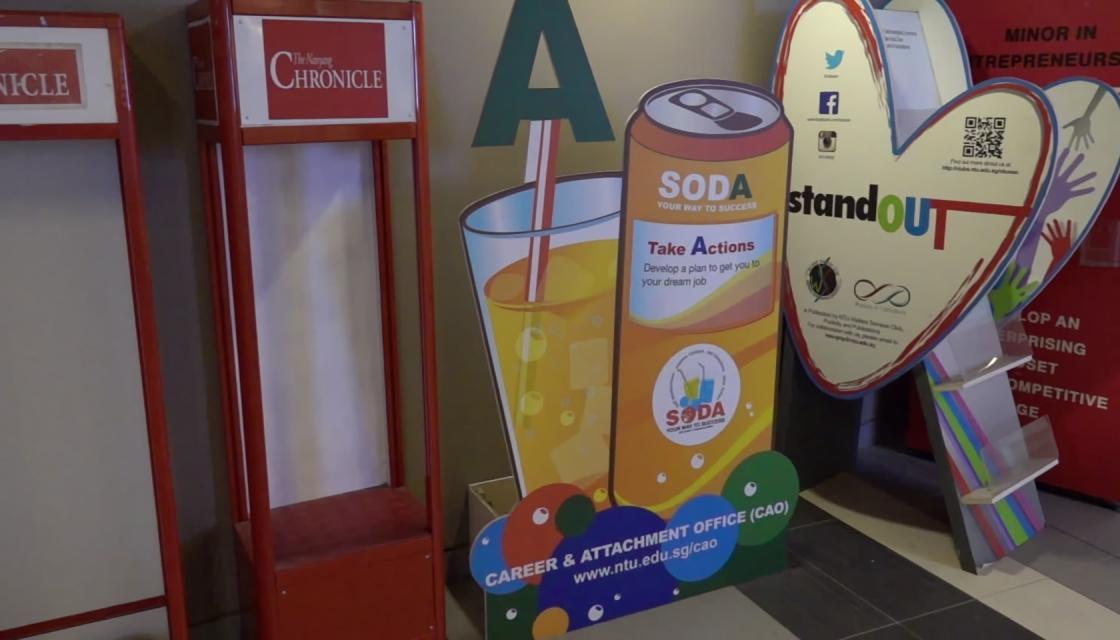} 
						\\
						 Low-light\&Blurry Input \hspace{\g} &
						PromptIR~\cite{promptir} \hspace{\g} &
						DACLIP-UIR~\cite{daclip} \hspace{\g} &
						DiffUIR~\cite{diffuir}\hspace{\g} &
      					\textbf{LoRA-IR (Ours)} &
                            GT \hspace{\g}
						\\
					\end{tabular}
				\end{adjustbox}
			
	}

  	\renewcommand{\arraystretch}{1}
	\resizebox{1.00\linewidth}{!} {
			\renewcommand{\h}{0.143}
			\newcommand{\w}{0.22}
				\begin{adjustbox}{valign=t}
					\begin{tabular}{cccccc}
						\includegraphics[height=\h \textwidth, width=\w \textwidth]{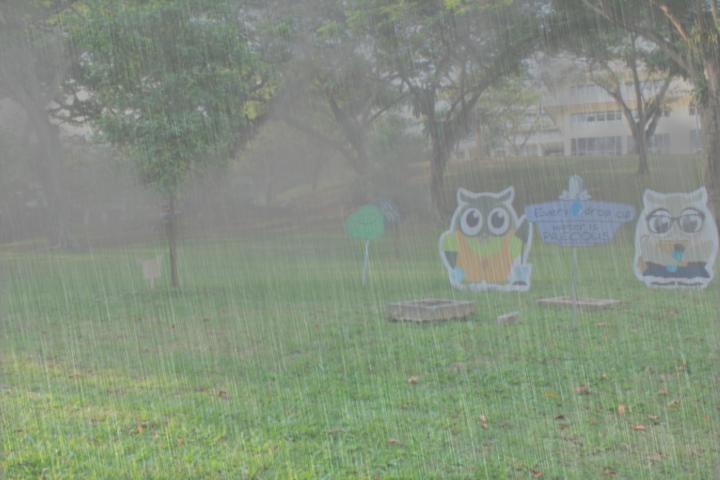} \hspace{\g} &
						\includegraphics[height=\h \textwidth, width=\w \textwidth]{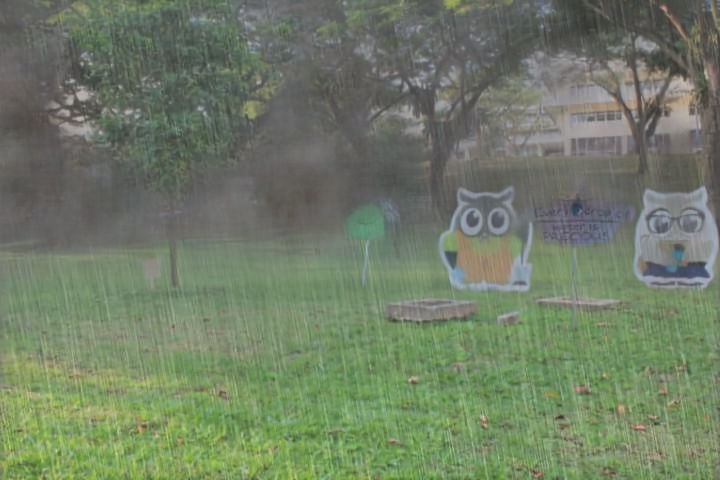} \hspace{\g} &
						\includegraphics[height=\h \textwidth, width=\w \textwidth]{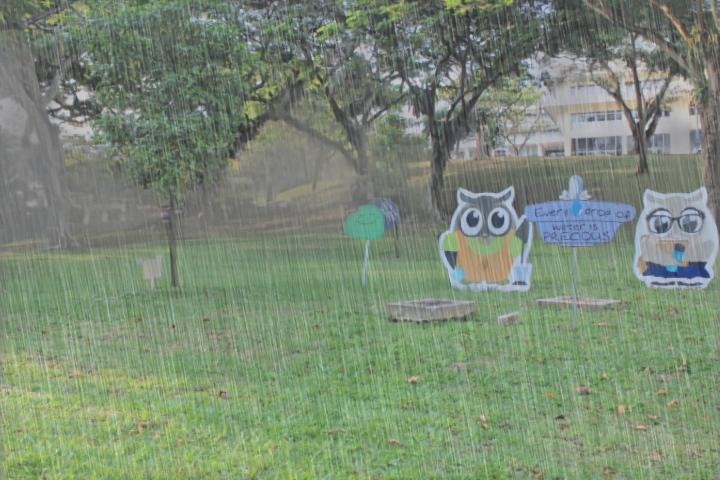} \hspace{\g} &
				\includegraphics[height=\h \textwidth, width=\w \textwidth]{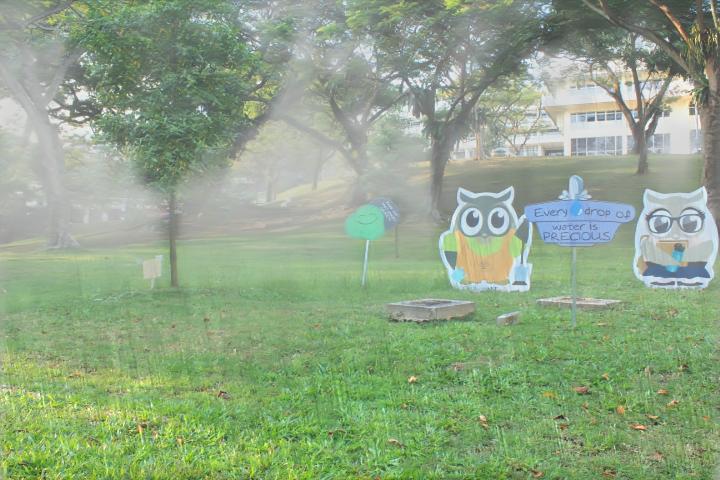} \hspace{\g} &
				\includegraphics[height=\h \textwidth, width=\w \textwidth]{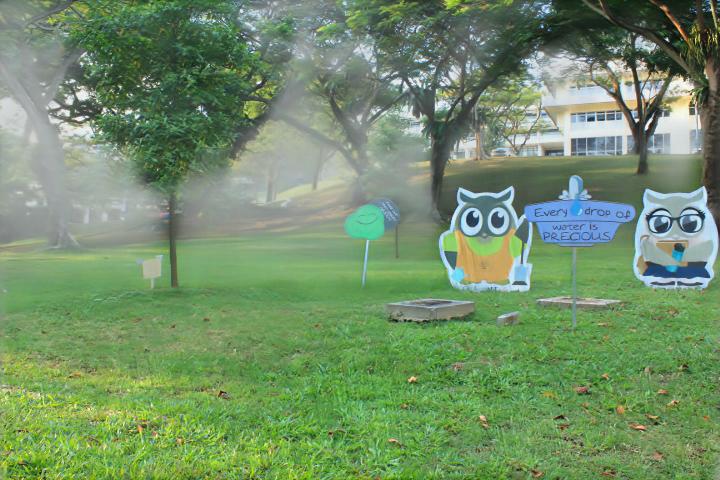} \hspace{\g} &
      			\includegraphics[height=\h \textwidth, width=\w \textwidth]{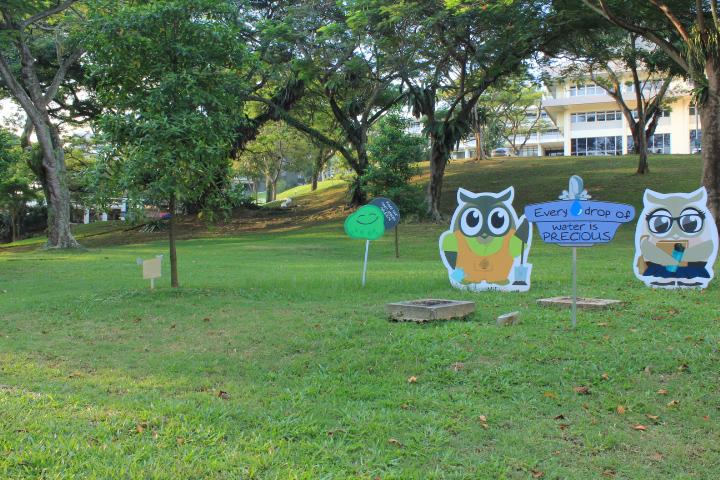} 
						\\
						 Rainy\&Hazy Input \hspace{\g} &
						 PromptIR~\cite{promptir} \hspace{\g} &
						DACLIP-UIR~\cite{daclip} \hspace{\g} &
						DiffUIR~\cite{diffuir}\hspace{\g} &
      					\textbf{LoRA-IR (Ours)} &
                            GT \hspace{\g}
						\\
					\end{tabular}
				\end{adjustbox}
			
	}

 \vspace{-0.3cm}
	\caption{Visual comparisons on challenging \textbf{\textit{mixed-degradation benchmarks}}. Please zoom in for a better view.}
	\label{fig:visual_mixed}
 \vspace{-0.5cm}
\end{figure*}

\noindent \textbf{Setting \Rmnum{5}.} Tab.~\ref{tab:de_predict} shows that, compared to DA-CLIP~\cite{daclip}, our DG-Router requires significantly fewer (approximately $64\times$) training parameters and a shorter (about $4\times$) training time, while achieving more accurate degradation predictions.  As shown in Tab.~\ref{tab:de10}, our LoRA-IR outperforms all compared general IR and all-in-one models in both distortion and perceptual metrics, showcasing the superiority of LoRA-IR. The detailed results for each task are provided in the \textbf{Appendix} due to the page limit.
\vspace{1mm}

\noindent \textbf{Mixed-degradataion Removal.} Considering that images in real-world scenarios may not contain only a single type of degradation, we further evaluate different all-in-one methods on mixed-degradation benchmarks. Our experiments include three mixed-degradation benchmarks: rain\&haze~\cite{outdoor}, low-light\&blur~\cite{lolblur}, and blur\&JPEG~\cite{REDS}. Tab.~\ref{tab:mixed} shows that our method has a significant advantage in handling challenging mixed-degradation scenarios. We provide visual results in Fig.~\ref{fig:visual_mixed}, showcasing the effectiveness of our method in handling mixed degradations.

\begin{table}[t]
\captionsetup{font=small}%
\scriptsize
\center
\caption{\textbf{[Setting \Rmnum{5}]} Comparison with DA-CLIP~\cite{daclip} on degradation prediction. Training time is evaluated using A100 GPU hours.}
\vspace{-0.2cm}
\begin{center}
\resizebox{\linewidth}{!}{
\setlength\tabcolsep{2.4pt}
\begin{tabular}{l|cc|ccccccccccc}
\toprule[0.15em]
\textbf{Method} & \makecell{\textbf{Trainable}\\ \textbf{Params}} &\makecell{\textbf{Training}\\ \textbf{Time}}  & Blur&Haze&JPEG&LL&Noise&RD&Rain&Shadow&Snow &Inpaint
         \\
         \midrule
         DA-CLIP &94.94M &12& 91.6 & 100 & 100 & 100 & 100 & 100 & 100 & 100 & 100 & 100
         \\
        \rowcolor{color4}
         \textbf{DG-Router} & \textbf{1.48M} & \textbf{2.67} &  100 & 100 & 100 &100 &100 &100 & 100 & 100 & 100 & 100
         \\
\bottomrule[0.15em]
\end{tabular}}
\end{center}
\vspace{-0.25cm}
\label{tab:de_predict}
\end{table}

\begin{table}[t] 
    \centering
        \caption{\textbf{[Setting \Rmnum{5}]} Quantitative comparisons for \textbf{\textit{10-task image restoration}}. Our LoRA-IR outperforms SOTA methods in both distortion and perceptual metrics.}
        \vspace{-0.2cm}
        \label{tab:de10}
\scalebox{0.8}{
\setlength{\tabcolsep}{10pt}
    \begin{tabular}{lcccc}
        \toprule[0.15em]
        \multirow{2}{*}{\textbf{Method}} &  \multicolumn{2}{c}{\textbf{Distortion}} & \multicolumn{2}{c}{\textbf{Perceptual}}  \\ \cmidrule(lr){2-3} \cmidrule(lr){4-5}
        &  PSNR $\uparrow$ & SSIM $\uparrow$ & LPIPS $\downarrow$ & FID $\downarrow$    \\
        \midrule
        NAFNet~\cite{nafnet}  & 26.34 & 0.847 & 0.159 & 55.68   \\
        Restormer~\cite{restormer}  & 26.43 & 0.850 & 0.157 & 54.03   \\ 
        IR-SDE~\cite{irsde}  & 23.64 & 0.754 & 0.167 & 49.18 \\
        \midrule
        AirNet~\cite{airnet}  & 25.62 & 0.844 & 0.182 & 64.86   \\
        PromptIR~\cite{promptir}  & \underline{27.14} & \underline{0.859} & 0.147 & 48.26   \\
        DACLIP-UIR~\cite{daclip}  & 27.01 & 0.794 & \underline{0.127} & \underline{34.89} \\
        \rowcolor{color4}
        \textbf{LoRA-IR}  & \textbf{28.64} & \textbf{0.878}& \textbf{0.118} & \textbf{34.26} \\
        \bottomrule[0.15em]
        \end{tabular}}
        \vspace{-0.4cm}
\end{table}

\begin{table}[t]
    \centering
    \caption{Quantitative comparison with SOTA all-in-one methods on \textbf{\textit{mixed-degradtaion benchmarks}}. Note that none of the models are trained on mixed-degradation data.}
    \vspace{-0.2cm}
    \setlength{\tabcolsep}{2.pt}
    \resizebox{0.50\textwidth}{!}{
        \begin{tabular}{ll|ccc|ccc|ccc}
        \toprule[0.15em]
        \multicolumn{2}{l|}{\multirow{2}{*}{\textbf{Method}}} &
        \multicolumn{3}{c|}{{\textbf{Rain \& Haze}~\cite{outdoor}}} & \multicolumn{3}{c|}{{\textbf{Low-light \& Blur}~\cite{lolblur}}} & \multicolumn{3}{c}{{\textbf{Blur \& JPEG}~\cite{REDS}}} \\
        \multicolumn{2}{c|}{}  & PSNR$\uparrow$ & SSIM$\uparrow$ & LPIPS$\downarrow$ & PSNR$\uparrow$ & SSIM$\uparrow$ & LPIPS$\downarrow$ & PSNR$\uparrow$ & SSIM$\uparrow$ & LPIPS$\downarrow$  \\
        \midrule
        \multicolumn{2}{l|}{AirNet~\cite{airnet}} & 14.02 & 0.627 & 0.477 & 13.84 & 0.611 & 0.344 & 22.79 & 0.692 & 0.389 \\
        \multicolumn{2}{l|}{PromptIR~\cite{promptir}} & 14.75 & 0.634 & 0.454 & 17.61 & 0.681 & 0.317 & 24.98 & 0.710 & 0.403\\
        \multicolumn{2}{l|}{DACLIP-UIR~\cite{daclip}} & 15.19 & 0.637 & 0.481 & 15.03 & 0.625 & 0.330 & 24.26 & 0.704&\underline{0.358} \\
        \multicolumn{2}{l|}{DiffUIR~\cite{diffuir}} & \underline{14.87}   & \underline{0.631}     & 0.459  & 15.97 & 0.657 & 0.339 & 24.86& 0.714 & \textbf{0.332} \\
        \rowcolor{color4}
        \multicolumn{2}{l|}{\textbf{LoRA-IR}} & 
    \textbf{15.41}    & \textbf{0.642}  & \textbf{0.445} & \textbf{20.59}
& \textbf{0.719} &  \textbf{0.305} & \textbf{25.05} & \textbf{0.716} & 0.359 \\
        \bottomrule[0.15em]
    \end{tabular}{}
    }
        \vspace{-0.3cm}
    \label{tab:mixed}
\end{table}

\begin{table}[t]
\captionsetup{font=small}%
\scriptsize
\center
\caption{Ablations of LoRA-IR on the AllWeather~\cite{valanarasu2022transweather} and mixed-degradation benchmarks (Mixed$_1$: low-light\&blur~\cite{lolblur}, Mixed$_2$: blur\&jpeg~\cite{REDS}).} 
\vspace{-0.2cm}
\setlength\tabcolsep{2pt}
\begin{center}
\resizebox{1.0\linewidth}{!}{
\begin{tabular}{l|cc|cc|cc|cc|cc}
    \toprule[0.15em]
   \multicolumn{1}{l|}{\multirow{2}{*}{\textbf{LoRA-IR}}} & \multicolumn{2}{c|}{\textbf{Snow}~\cite{snow100k}} & \multicolumn{2}{c|}{\textbf{Rain}~\cite{outdoor}}& \multicolumn{2}{c|}{\textbf{Raindrop}~\cite{raindrop}} &\multicolumn{2}{c|}{\textbf{Mixed}$_1$\cite{lolblur}} & \multicolumn{2}{c}{\textbf{Mixed}$_2$\cite{REDS}} \\
          & PSNR  & SSIM   & PSNR  & SSIM  & PSNR  & SSIM  & PSNR  &SSIM  &PSNR  & SSIM  
         \\ \midrule
                 \rowcolor{color4}
         \textbf{DG-Router} & 32.28 & 0.930 & 32.62 & 0.945 & 33.39 & 0.949 & 20.59 & 0.719 & 25.05 & 0.716  \\
        w/o high reso. & 32.19 & 0.929 & 32.31 & 0.942 & 33.20 & 0.947 & 19.51 & 0.711 & 24.94 & 0.714  \\
        \midrule
                \rowcolor{color4}
         \textbf{DAM} & 32.28 & 0.930 & 32.62 & 0.945 & 33.39 & 0.949 & 20.59 & 0.719 & 25.05 & 0.716  \\  
         w/o DAM & 32.07 & 0.927 & 32.28 & 0.941 & 33.12 & 0.945 & 18.91 & 0.704 & 24.89 & 0.711  \\  
         AdaLN~\cite{perez2018film} Modulator & 32.13 & 0.926 & 32.33 & 0.938 & 33.22 & 0.942 & 18.44 & 0.700 & 24.77 & 0.705  \\ 
         \midrule
            \rowcolor{color4}
         \textbf{Mixture of LoRA Expert} & 32.28 & 0.930 & 32.62 & 0.945 & 33.39 & 0.949 & 20.59 & 0.719 & 25.05 & 0.716  \\ 
         w/o LoRA Expert & 32.01 & 0.925 & 32.19 & 0.938 & 33.03 & 0.944 & 16.79 & 0.675 & 24.55 & 0.709  \\ 
         \bottomrule[0.15em]
    \end{tabular}}
\end{center}
    \label{tab:ablation_model}
        \vspace{-0.3cm}

\end{table}

\subsection{Ablation Study}
We perform ablation studies to examine the role of each component in our proposed LoRA-IR. To comprehensively validate our method, we conduct experiments on the AllWeather~\cite{valanarasu2022transweather} and mixed-degradation~\cite{lolblur,REDS} benchmarks. In Tab.~\ref{tab:ablation_model}, we start with LoRA-IR and systematically remove or replace modules, including the high-resolution techniques in DG-Router, the DAM module (we also attempt to use AdaLN~\cite{perez2018film} for feature modulation), and the mixture of LoRA expert design. We find that LoRA-IR consistently outperforms its ablated versions across all benchmarks, highlighting the critical importance of these components. Notably, our mixture of LoRA expert design significantly improves the model's performance on mixed-degradation benchmarks, enhancing the model's generalizability in real-world scenarios. More detailed ablations, model efficiency comparison and analysis are provided in the \textbf{Appendix}.

\section{Conclusion}
This paper introduces LoRA-IR, a flexible framework that dynamically leverages compact low-rank experts to facilitate efficient all-in-one image restoration. We propose a CLIP-based Degradation-Guided Router (DG-Router) to extract robust degradation representations, requiring minimal training parameters and time. With the valuable guidance of the DG-Router, LoRA-IR dynamically integrates different low-rank experts, enhancing architectural adaptability while preserving computational efficiency. Across 14 image restoration 
tasks and 29 benchmarks, LoRA-IR demonstrates its state-of-the-art performance and strong generalizability.

{
    \small
    \bibliographystyle{ieeenat_fullname}
    \bibliography{main}
}


\end{document}